\documentclass[lettersize,journal]{IEEEtran}
\usepackage{amsmath,amsfonts}
\usepackage{algorithmic}
\usepackage{algorithm}
\usepackage{array}
\usepackage[caption=false,font=normalsize,labelfont=sf,textfont=sf]{subfig}
\usepackage{textcomp}
\usepackage{stfloats}
\usepackage{url}
\usepackage{verbatim}
\usepackage{graphicx}
\usepackage{cite}
\usepackage{subfig}
\usepackage{booktabs}
\usepackage{multirow}
\usepackage{rotating}
\usepackage{caption}
\usepackage{booktabs}
\usepackage{geometry}
\usepackage{amsmath}
\usepackage{subcaption}
\usepackage[skip=0.3em, justification=centering]{caption}
\usepackage[numbers,sort&compress]{natbib}
\setcitestyle{numbers,square}
\begin{document}

\title{WeatherSeg: Weather-Robust Image Segmentation using Teacher-Student Dual Learning and Classifier-Updating Attention}

\author{
\IEEEauthorblockN{Zhang Zhang}
\thanks{Zhang Zhang is with the School of Artificial Intelligence and Information Engineering, Zhejiang University of Science and Technology, Hangzhou, China (email: 123138@zust.edu.cn).}
\IEEEauthorblockN{Yifeng Zeng}
\thanks{Yifeng Zeng is with the Department of Computer and Information Sciences, Northumbria University, Newcastle, United Kingdom (email: Yifeng.Zeng@northumbria.ac.uk).}
\IEEEauthorblockN{Houshi Jiang}
\thanks{Houshi Jiang is with Beijing Quinovare Medical Technology Co., Ltd, Beijing, China (email: jianghoushi@quinovare.com).}
\IEEEauthorblockN{Yinghui Pan*}
\thanks{Yinghui Pan is the corresponding author. She is with the School of Artificial Intelligence, Shenzhen University, Shenzhen, China (email: panyinghui@szu.edu.cn).}
}

\maketitle

\begin{abstract}
\textbf{WeatherSeg}, an advanced semi-supervised segmentation framework, addresses autonomous driving's environmental perception challenges in adverse weather while reducing annotation costs. This framework integrates a \textbf{Dual Teacher-Student Weight-Sharing Model (DTSWSM)} that enables knowledge distillation from weather-affected images, and a \textbf{Classifier Weight Updating Attention Mechanism (CWUAM)} that dynamically adjusts classifier weights based on environmental attributes. Comprehensive evaluations demonstrate that \textbf{WeatherSeg} significantly outperforms baseline models in both accuracy and robustness across various weather conditions, including clear, rainy, cloudy, and foggy scenarios, establishing it as an effective solution for all-weather semantic segmentation in autonomous driving and related applications.

\end{abstract}


\section{Introduction}
\label{Intro}

\IEEEPARstart{S}{emantic} Segmentation \cite{HAO2020302}, \cite{MO2022626}, a cornerstone of computer vision, enables the pixel-wise classification of images into distinct categories. Recent deep learning breakthroughs have dramatically improved performance in benchmark datasets. This capability critically supports autonomous driving by enabling the comprehension of the road scene, accurate object detection, and reliable obstacle avoidance. However, real-world driving environments present substantial challenges due to constantly varying lighting and weather conditions. This environmental complexity makes comprehensive data annotation exceptionally difficult, severely limiting the generalisation capability of current perception models.

Extreme weather conditions—including heavy rain, dense fog, and glaring sunlight—directly impair autonomous systems' perception reliability. These conditions not only obscure visual clarity but also hinder accurate environmental assessment and timely decision-making. Furthermore, obtaining large-scale, clearly annotated datasets covering diverse extreme weather scenarios requires prohibitive cost and time investment.

Many existing approaches attempt to enhance training sample diversity through data augmentation techniques. These methods seek to optimise sensor cost-effectiveness while improving model robustness across environments. Despite these efforts, autonomous vehicles still face a critical shortage of adequate training data for extreme weather operations. This shortage has shifted research focus toward developing lightweight, easily integrable frameworks for existing segmentation networks.

Semi-supervised semantic segmentation (SSL)\cite{HAO2020302} networks offer a promising solution by reducing dependence on expensively labeled data. These networks leverage semi-supervised learning to improve data processing efficiency and enhance visual interpretation under adverse conditions. SSL techniques effectively balance abundant unlabelled data utilisation with limited labeled sample precision.

Nevertheless, existing SSL methods face persistent challenges in autonomous driving applications. Under severe weather conditions where visual information becomes heavily obscured, most methods struggle to extract sufficiently informative features from unlabelled data. Inconsistent loss calculations between labeled and unlabelled data further degrade practical deployment performance. The continued reliance on pixel-level annotations creates a fundamental bottleneck considering the extensive weather and lighting variations in real-world driving. These limitations underscore the urgent need for computationally efficient approaches that concurrently reduce camera and sensor system costs.

A clear gap exists in current literature: existing methods fail to extract robust features from images degraded by adverse weather and struggle with inconsistent loss calculations between labeled and unlabelled data. These shortcomings hinder the development of reliable perception systems for autonomous driving \cite{9913352}.


To overcome these critical challenges, we present \textbf{WeatherSeg}—a novel universal framework for semi-supervised semantic segmentation. Our lightweight architecture seamlessly integrates with existing networks, effectively processes blurred adverse weather imagery, intelligently assigns weights, and resolves common mean-teacher training limitations. \textbf{WeatherSeg} provides a flexible, efficient, and robust environmental perception solution that significantly improves model generalisation, simplifies training complexity, enhances stability, and reduces hardware costs. By efficiently leveraging unlabelled data while minimising dependency on expensive annotations, our framework streamlines training processes and achieves substantial cost reduction.

\begin{figure*}[htbp!]
        \centering
        \includegraphics[width=12cm,height=3cm]{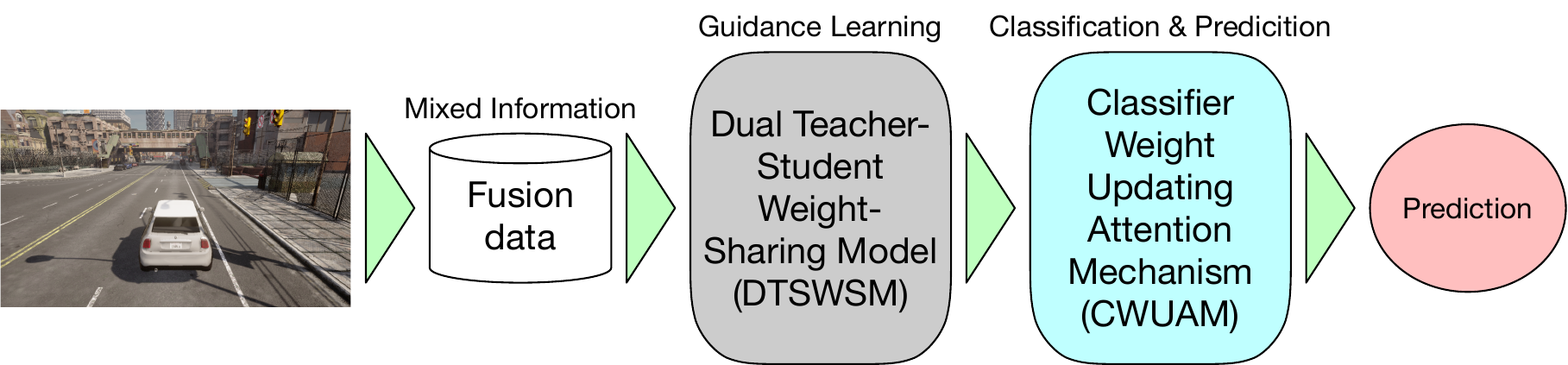}
        \caption{WeatherSeg: The whole network architecture of the dual teacher-student with classifier weight transformer. This architecture comprises two main components. The first is the Dual Teacher-Student Weight-Sharing Model (DTSWSM), which mixes and fuses dynamic vehicle data. It guides the training of shared parameters and distills primary feature information. The second component is the Classifier Weight Updating Attention Mechanism (CWUAM). Centered around a classifier attention mechanism, \emph{CWUAM} sharpens the output from the first component to achieve superior semi-supervised semantic segmentation results.}
        \label{fig:architecture}
\end{figure*}

Our work makes three primary contributions as Fig.\ref{fig:architecture}:

\begin{itemize}
\item \textbf{Dual Teacher-Student Weight-Sharing Model (DTSWSM)}: Processes fused labeled and unlabelled data streams using consistency loss training and classification pre-training. This dual-channel architecture employs separate training paths with weight sharing, enabling teacher models to effectively guide student networks and substantially improve prediction performance.

\item \textbf{Classifier Weight Updating Attention Mechanism (CWUAM)}: Sharpens distillation model predictions by updating classifier weights using preceding model outputs, achieving strong performance on unlabelled data.

\item \textbf{Comprehensive Evaluation}: Demonstrates through extensive testing across diverse conditions—including clear weather, varying fog/rain intensities, and day-night transitions—that \textbf{WeatherSeg} effectively handles complex visual information while significantly improving foundational model generality and adaptability.
\end{itemize}
The paper organises as follows: Section \ref{liter} details existing semantic segmentation methods, Section \ref{model} comprehensively describes our proposed framework and training methodology, Section \ref{exp} presents comparative experimental results, and Section \ref{con} summarises our work while outlining future directions.

\section{Related Work}
\label{liter}
Semantic segmentation is a cornerstone technology for autonomous vehicles, enabling them to achieve a detailed, pixel-level understanding of their environment \cite{sellat2022semantic}, \cite{app11198802}. While foundational deep learning models have advanced this field, significant challenges remain, particularly in label-scarce and adverse weather scenarios. This section analyses existing architectures and semi-supervised techniques, highlighting the critical gaps our proposed framework addresses.

Current methodologies shaping semi-supervised segmentation include \cite{HOU2024110089}, \cite{Liu_2022_CVPR}, \cite{2266853},  \cite{Ouali_Hudelot_Tami_2020}, pseudo-labelling methods \cite{Wang_2022_CVPR}, \cite{WANG2022108925}, \cite{Liu_2022_CVPR}, and entropy minimisation techniques \cite{9428304}. Recent advances also explore generative modelling\cite{li2021semantic}. Techniques like View-coherent Correlation Consistency\cite{HOU2024110089} maintain prediction robustness across augmented image views, while Cross-Consistency Training\cite{Ouali_Hudelot_Tami_2020} introduces perturbations to enforce decoder consistency. Other methods\cite{Liu_2022_CVPR} combine Mean Teacher networks\cite{tarvainen2017mean} with confidence-weighted loss to refine consistency learning and reduce prediction uncertainty. Adaptive threshold adjustment\cite{Wang_2022_CVPR}, \cite{Liu_2022_CVPR} further improves pseudo-labelling quality throughout training.

Pioneering architectures like Fully Convolutional Network (FCN) \cite{101145}, SegNet \cite{badrinarayanan2017segnet}, U-Net \cite{gautam2022image}, \cite{SONG2024122406}, and DeepLab \cite{chen2017deeplab} established the basis for end-to-end semantic segmentation. FCN introduced pixel-wise prediction but suffers from high computational costs. SegNet and U-Net utilise encoder-decoder structures to refine spatial details, yet they still lose critical spatial information, struggle with occlusions, and demand significant computational resources. Researchers have attempted to optimise these models through compression and hardware acceleration \cite{QIU2023109383}, \cite{SAMBATURU2023109011}. However, incremental improvements often introduce new limitations. For example, Gautam et al. \cite{gautam2022image} enhanced U-Net with clustering algorithms but increased computational overhead and manual effort. Similarly, Lu et al. \cite{lu2021simpler} developed an effective few-shot segmentation method that still requires validation for robust performance in challenging weather conditions that affect entire datasets.

To mitigate the intensive labor and cost of data annotation, researchers have turned to semi-supervised learning (SSL) \cite{2266853}, \cite{HOU2024110089}, \cite{LU2024125456}. Common SSL strategies include \textbf{consistency training}, \textbf{pseudo-labelling}, and \textbf{Mean Teacher-based models}. Consistency training methods \cite{2266853}, \cite{HOU2024110089}, \cite{LU2024125456}, \cite{xu2022semi}, \cite{xin2024enhancing} enforce stable predictions under perturbations but increase computational load and are sensitive to data distribution bias. Further optimisations using advanced data augmentation \cite{hoyer2023improving}, \cite{yi2021learning} only add to this complexity. Pseudo-labelling techniques \cite{wang2020semi}, \cite{Wang_2022_CVPR}, \cite{xu2023ambiguity}, \cite{WANG2022108925} leverage unlabelled data but remain highly sensitive to hyper-parameters and struggle to generalise from noisy initial labels.

The \textbf{Mean Teacher model} \cite{tarvainen2017mean} offers an elegant synthesis of these ideas, yet it complicates the training process and remains vulnerable to data distribution shifts. Subsequent enhancements, such as dual attention mechanisms \cite{xu2021dual}, \cite{cui2019semi} and cross-teacher modules \cite{xiao2022semi}, \cite{jin2022semi}, improve accuracy at the cost of greater training complexity, hyper-parameter sensitivity, and a higher risk of overfitting, which ultimately reduces model interpretability and real-world feasibility. Other approaches that distill features from 2D networks to enhance LiDAR segmentation \cite{8691698}, \cite{WANG202020} show promise but depend heavily on synthetic data quality and have yet to solve generalisation issues.

\section{Proposed WeatherSeg}
\label{model}
\subsection{Objective}
\begin{figure}[htbp!]
        \centering
        \includegraphics[width=8cm,height=8cm]{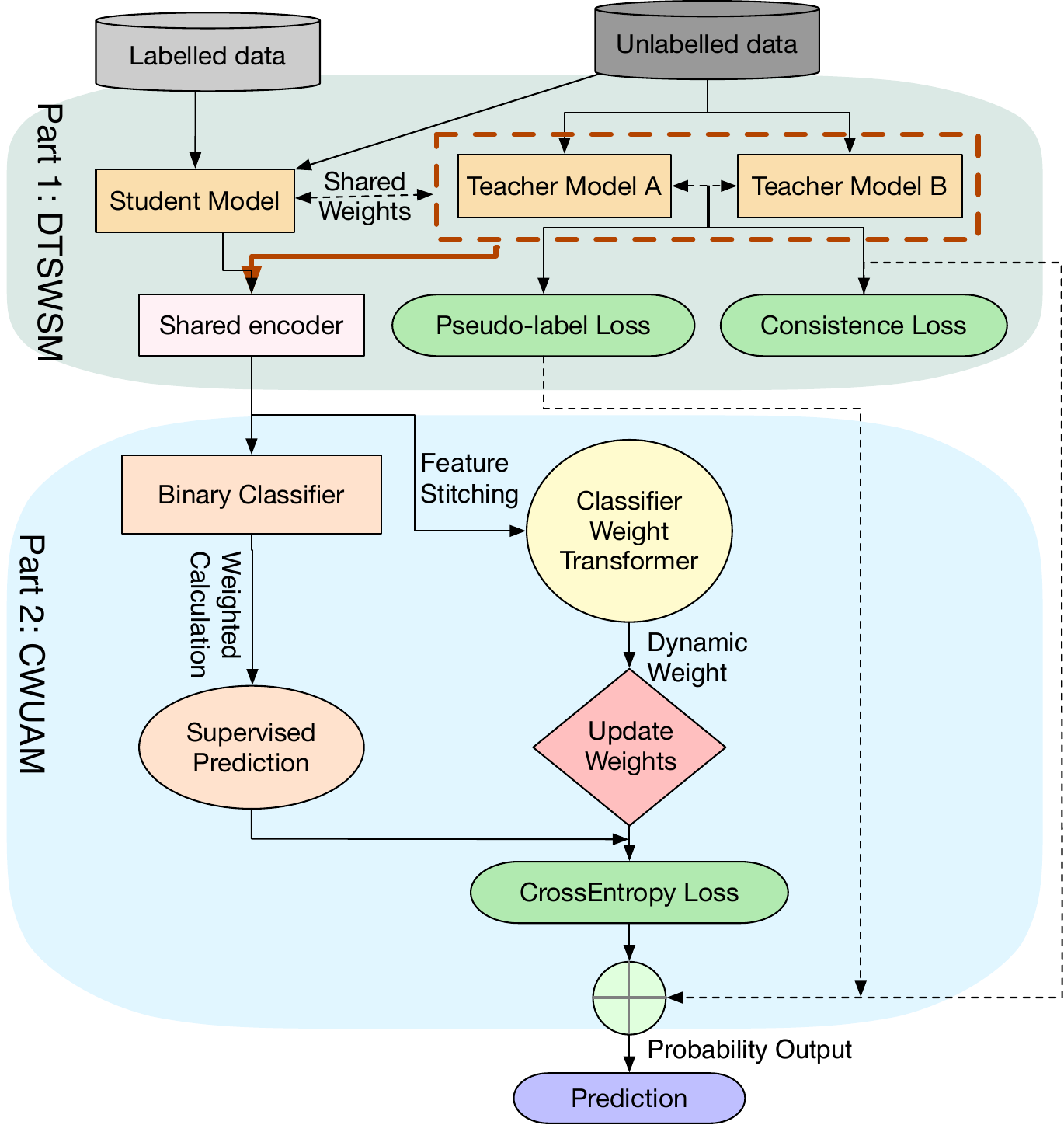}
        \caption{WeatherSeg: The first part is DTSWSM, whose primary role is to provide a stable foundational feature representation for the subsequent Classifier Weight Updating Attention Mechanism (\emph{CWUAM}) dynamic weighting through parameter sharing and feature unification. \emph{CWUAM} sharpens the basic feature output of the first part, accurately classifies the category importance, and represents the stable features as task-adaptive prediction results.}
        \label{fig:architecture-1}
\end{figure}

We present \textbf{WeatherSeg}, an innovative semi-supervised semantic segmentation framework designed for autonomous driving in adverse weather (Fig. \ref{fig:architecture-1}). Our architecture synergistically integrates a \textbf{Dual Teacher-Student Weight-Sharing Model (\emph{DTSWSM})}with a \textbf{Classifier Weight Updating Attention Mechanism (\emph{CWUAM})}. This design effectively leverages both limited labeled and abundant unlabelled data to achieve robust environmental perception.

\subsection{Problem Definition}
        We define our labeled dataset as $\mathcal{D}_L = \{(x_i, y_i)\}_{i=1}^N$ and our unlabelled dataset as $\mathcal{D}_U = \{x_j\}_{j=1}^M$. Our architecture consists of a shared encoder $E_\theta(\cdot)$, a student classification header $S_\omega(\cdot)$, dual teacher classification headers $T^{(1)}_\phi(\cdot)$ and $T^{(2)}_\psi(\cdot)$, and a Classifier Weight Transformer $\text{CWT}_\eta(\cdot)$.
        
\subsection{Architecture and Analysis}
\subsubsection{Dual Teacher-Student Weight-Sharing Model(DTSWSM)}
\begin{figure*}[htbp!]
                \centering
                \includegraphics[width=12cm,height=5cm]{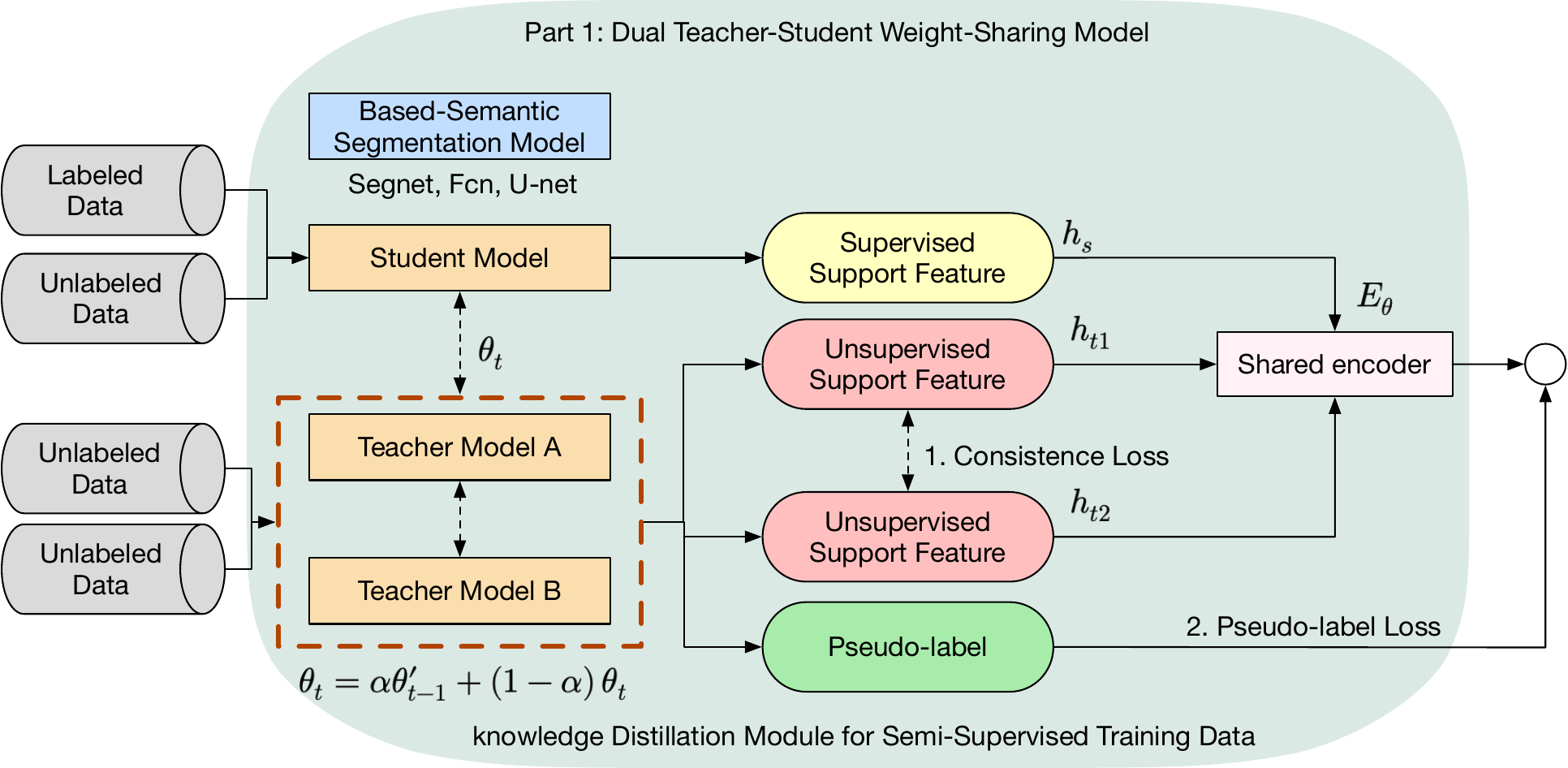}
                \caption{A. Dual Teacher-Student Weight-Sharing Model(DTSWSM): It uses a shared encoder to extract unified features from mixed data. A supervised student model and dual teacher models (updated via independent EMAs) train concurrently, with a consensus mechanism generating pseudo-labels from high-confidence samples.}
                \label{fig:part1}
        \end{figure*}

        The \emph{DTSWSM} (Fig. \ref{fig:part1}) is our foundational component for robust feature learning and pseudo-label generation. It processes both labeled and unlabelled data through student and dual-teacher pathways, using consistency-based training to ensure a stable feature representation.
        
        We selected a Transformer-based attention mechanism for the \emph{CWUAM} because weather distortions are complex and context-dependent. Unlike a static MLP, an attention mechanism models these dynamic, non-linear relationships by allowing the student’s features to query relevant information from the teachers’ stable features on a per-pixel basis, providing a more powerful and adaptive re-weighting capability.
        
\textbf{Shared Encoder Feature Extraction}: A shared encoder $E_\theta(\cdot)$ processes both labeled and unlabelled inputs to extract unified feature representations $h = E_\theta(x)$. For a batch of size $B$, this yields student features $h_s \in \mathbb{R}^{B \times D}$ and teacher features $h_{t1}, h_{t2} \in \mathbb{R}^{B \times D}$.

\textbf{Teacher Model Parameter Update (Exponential Moving Average - EMA)}: We update the teacher models using an Exponential Moving Average (EMA) to ensure stable parameter evolution without direct gradient updates. The update rule for teacher $i \in \{1,2\}$ at time step $t$ is Equ.\ref{equ:ema}:
        \begin{equation}
        \label{equ:ema}
        \theta_{T_i}^{(t)} = \alpha \theta_{T_i}^{(t-1)} + (1-\alpha) \theta_{S}^{(t)}
        \end{equation}
        where $\theta_{T_i}$ and $\theta_{S}$ are the teacher and student parameters, and $\alpha$ is the smoothing coefficient (typically $0.99$).
        
        \emph{Derivation \& Advantage}: This EMA update bounds the change in teacher parameters relative to the student's, as shown in Equ.\ref{equ:ema_bound}, ensuring model stability and providing more consistent pseudo-labels.
        \begin{equation}
        \label{equ:ema_bound}
        | \theta_{T}^{(t+1)} - \theta_{T}^{(t)} | = (1-\alpha) | \theta_{S}^{(t+1)} - \theta_{T}^{(t)} | \approx (1-\alpha) | \theta_{S}^{(t+1)} - \theta_{S}^{(t)} |
        \end{equation}
        
\textbf{Consistency Loss ($\mathcal{L}_{Consist}$)}: We enforce prediction consistency between the two independently maintained teacher models to enhance feature robustness as Equ.\ref{equ:LC}:
        \begin{equation}
        \label{equ:LC}
        \mathcal{L}_{\text{Consist}} = \frac{1}{B_U} \sum_{i=1}^{B_U} \| T^{(1)}_\phi(E_\theta(x_i)) - T^{(2)}_\psi(E_\theta(x_i)) \|_2^2
        \end{equation}
        
        \emph{Derivation \& Advantage}: Minimising this MSE loss aligns the outputs of $T^{(1)}_\phi$ and $T^{(2)}_\psi$, which improves feature discriminability and the reliability of pseudo-labels generated via teacher consensus.
        
\textbf{Teacher Consensus Pseudo-Label Generation}: Our dual-teacher consensus mechanism generates high-quality pseudo-labels by averaging the teachers' softmax probabilities as Equ.\ref{equ:pavg}:

        \begin{equation}
        \label{equ:pavg}
        p_{\text{avg},j} = \frac{1}{2} \left( \text{Softmax}(T^{(1)}_\phi(E_\theta(x_j))) + \text{Softmax}(T^{(2)}_\psi(E_\theta(x_j))) \right)
        \end{equation}
        
        We then assign a pseudo-label $\tilde{y}_j$ only if the maximum confidence exceeds a threshold $\tau=0.95$ as Equ.\ref{equ:tilde}:
        
        \begin{equation}
        \label{equ:tilde}
        \tilde{y}_{j} = \begin{cases} \arg\max_{c}(p_{\text{avg},j})_{c} & \text{if } \max_{c}(p_{\text{avg},j}) > \tau \\ \text{ignore} & \text{otherwise} \end{cases}
        \end{equation}
        
        \emph{Derivation \& Advantage}: Averaging predictions ($p_{t1}, p_{t2}$) from the two EMA-stabilised teachers reduces pseudo-label variance. The variance of the average prediction as Equ.\ref{equ:var}:
        \begin{equation}
        \label{equ:var}
        \text{Var}(p_{\text{avg}}) = \frac{1}{4}(\text{Var}(p_{t1})+\text{Var}(p_{t2})+2\text{Cov}(p_{t1},p_{t2}))
        \end{equation}
        
        Assuming low correlation between teachers and that $\text{Var}(p_{t1}) \approx \text{Var}(p_{t2}) \approx \text{Var}(p_{\text{single}})$, we achieve as Equ.\ref{equ:var-reduct}:
        \begin{equation}
        \label{equ:var-reduct}
        \text{Var}(p_{\text{avg}}) \approx \frac{1}{2}\text{Var}(p_{\text{single}})
        \end{equation}
        
        This variance reduction of approximately 50$\%$ yields higher-quality pseudo-labels and more effective learning from unlabeled data.
        
\textbf{Pseudo-Label Loss ($\mathcal{L}_{PL}$)}: The student model trains on high-confidence pseudo-labels using a dynamically weighted cross-entropy loss, allowing it to learn from the unlabeled dataset as Equ.\ref{equ:LPL}:
\begin{equation}
\label{equ:LPL}
\begin{split}
\mathcal{L}_{\text{PL}} = -\frac{1}{|\mathcal{M}|} \sum_{j \in \mathcal{M}} \sum_{c=1}^C \mathbb{I}(\tilde{y}_j=c) \\
\times \log\left( \text{Softmax}(S_\omega(E_\theta(x_j)) \odot w_j)_c \right)
\end{split}
\end{equation}
        Here, $\mathcal{M}$ is the set of high-confidence samples, and $w_j$ are dynamic weights from the \emph{CWUAM}.
        
\subsubsection{Classifier Weight Updating Attention Mechanism(CWUAM)}
\begin{figure*}[htbp!]
        \centering
        \includegraphics[width=12cm,height=5cm]{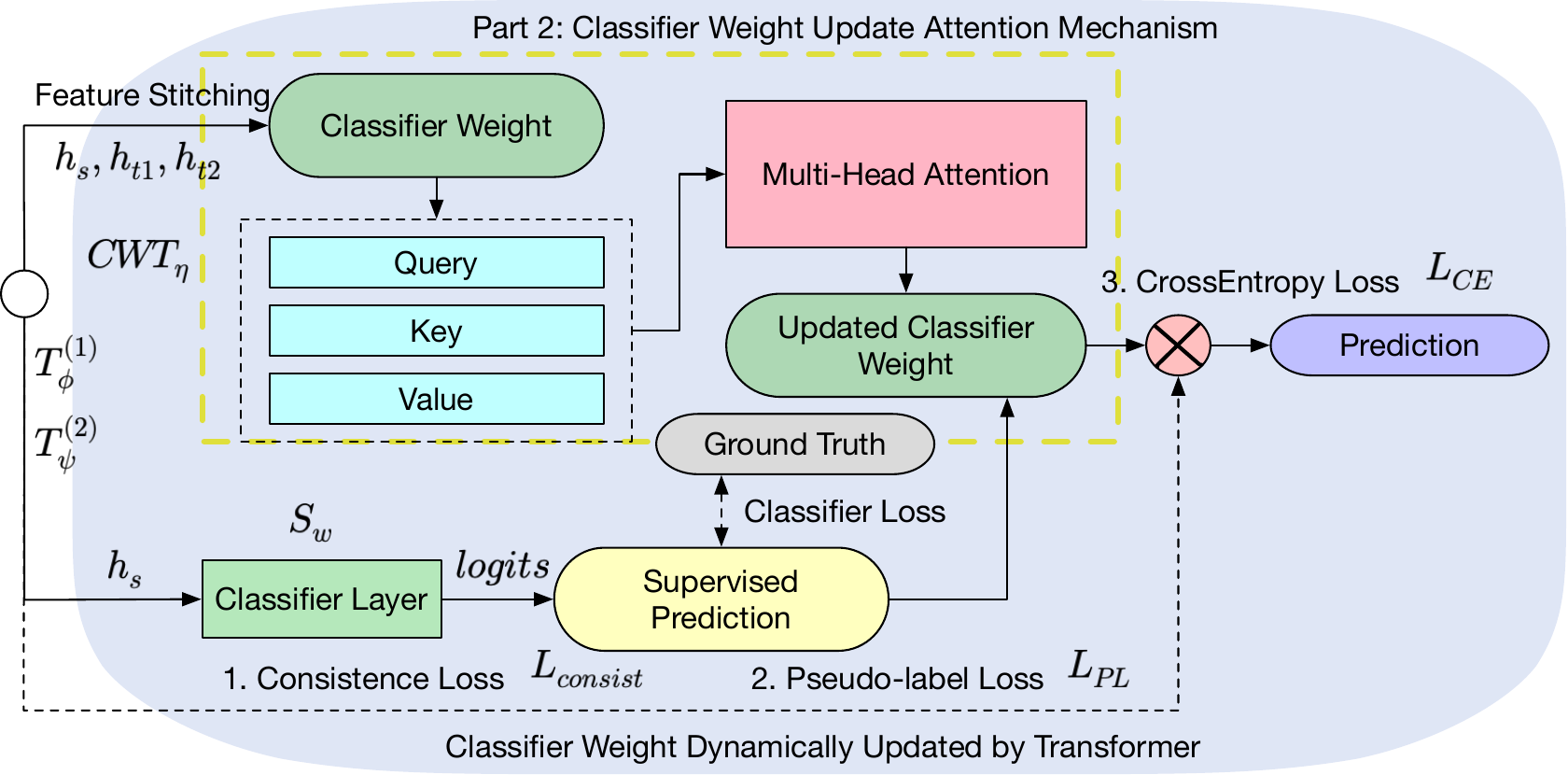}
        \caption{B. Classifier Weight Updating Attention Mechanism: By integrating the features of dual teachers and students through a dynamic attention mechanism, adaptive category weights for samples are generated.}
        \label{fig:part2}
\end{figure*}

        The \emph{CWUAM} (Fig. \ref{fig:part2}) is a lightweight, Transformer-based module that dynamically refines classification. It uses multi-head attention to generate sample-specific, task-adaptive category weights, which sharpens the model's focus on critical features and improves accuracy, especially for difficult samples and rare categories.
        
\textbf{Feature Concatenation}: We first concatenate the feature representations from the two teacher models ($h_{t1}, h_{t2}$) to form a comprehensive teacher feature vector $H_t \in \mathbb{R}^{B \times 2D}$, which provides a rich context for the attention mechanism as Equ.\ref{equ:ht}:
        \begin{equation}
        \label{equ:ht}
        H_t = \text{concat}(h_{t1}, h_{t2})
        \end{equation}
        
\textbf{Dynamic Weight Generation (CWT)}: The Classifier Weight Transformer ($\text{CWT}_\eta$) generates dynamic, class-specific weights $w_{\text{class}} \in \mathbb{R}^{B \times C}$ for each sample. It takes student features $h_s$ as the Query (Q) and concatenated teacher features $H_t$ as the Key (K) and Value (V) as Equ.\ref{equ:Q}:
        \begin{equation}
        \label{equ:Q}
        Q = h_s W_Q \quad K = H_t W_K \quad V = H_t W_V
        \end{equation}
        where $W_Q, W_K, W_V$ are learnable projection matrices. The attention mechanism computes a weighted sum of the values based on query-key similarity as Equ.\ref{equ:Attn}:
        \begin{equation}
        \label{equ:Attn}
        \text{Attention}(Q, K, V) = \text{Softmax}\left(\frac{QK^T}{\sqrt{d_k}}\right) V
        \end{equation}
        The attention output is then passed through a LayerNorm and an MLP to produce the final dynamic weights as Equ.\ref{equ:w_class}:
        \begin{equation}
        \label{equ:w_class}
        w_{\text{class}} = \text{MLP}(\text{LayerNorm}(\text{Attention}(Q,K,V)))
        \end{equation}
        
        \emph{Derivation \& Advantage}: The attention mechanism adaptively recalibrates teacher features based on the student's context, learning dynamic weights $w_c \propto \text{Attention}(h_s, H_t)_c$. This granular, per-pixel adjustment sharpens classification decisions in challenging scenarios.
        
\textbf{Supervised Cross-Entropy Loss ($\mathcal{L}_{CE}$)}: For labeled data, we modulate the student's logits with the dynamic weights $w_{\text{class}}$ before computing the final prediction as Equ.\ref{equ:p_final}:
        \begin{equation}
        \label{equ:p_final}
        p(y_i|x_i) = \text{Softmax}(S_\omega(E_\theta(x_i)) \odot w_{\text{class}}^{(i)})
        \end{equation}
        We then compute the supervised loss using the standard cross-entropy function as Equ.\ref{equ:LCE}:
        \begin{equation}
        \label{equ:LCE}
        \mathcal{L}_{\text{CE}} = -\frac{1}{B_L} \sum_{i=1}^{B_L} \sum_{c=1}^C y_{i,c} \log(p(y_i|x_i)_c)
        \end{equation}
        
\subsection{Results: Training Methodology and Theoretical Advantages}
        Our framework uses a joint optimisation strategy that combines supervised learning on labeled data with pseudo-labelling and consistency regularisation on unlabelled data to ensure efficient learning and robust generalisation.
        
\subsubsection{Optimisation Strategy}:
        We optimise the student model parameters by minimising a total loss function, which is a weighted sum of four components as Equ.\ref{equ:total}:
        \begin{equation}
        \begin{split}
        \label{equ:total}
        \mathcal{L}_{\text{total}} = \underbrace{\mathcal{L}_{\text{CE}}}_{\text{Supervised}} + \lambda_1 \underbrace{\mathcal{L}_{\text{PL}}}_{\text{Pseudo-label}} +\\ \lambda_2 \underbrace{\mathcal{L}_{\text{Consist}}}_{\text{Consistency}} + \lambda_3 \underbrace{\mathcal{L}_{\text{reg}}}_{\text{Weight Regularization}}
        \end{split}
        \end{equation}
        where $\mathcal{L}_{\text{reg}} = \|w_{\text{class}} - \mathbf{1}\|_2^2$ encourages the dynamic weights to stay close to unity. We empirically set $\lambda_1=0.3, \lambda_2=0.1, \lambda_3=0.01$. We update the parameters through three distinct pathways:
        
\begin{itemize}	
        \item \textbf{Student parameters ($\theta, \omega$)}: Updated via back-propagation on $\mathcal{L}_{\text{total}}$.
        \item \textbf{CWT Parameters ($\eta$)}: Updated via back-propagation on the combined supervised and pseudo-label losses ($\mathcal{L}_{\text{CE}}+\mathcal{L}_{\text{PL}}$).
        \item \textbf{Teacher parameters ($\phi, \psi$)}: Updated via the EMA rule in Equ. \ref{equ:ema} without back-propagation.
\end{itemize}

\subsubsection{Theoretical Advantages}:
        Our framework offers several key theoretical advantages:
        \begin{itemize}
        \item \textbf{Reduced Variance in Pseudo-Labels}: As derived in Equ. \ref{equ:var-reduct}, our dual-teacher consensus mechanism reduces pseudo-label variance by approximately 50$\%$ compared to a single-teacher model, yielding higher-quality supervisory signals.
        \item \textbf{Stability of EMA Updates}: The EMA updates (Equ.\ref{equ:ema}) ensure the teacher models evolve smoothly, providing stable and consistent targets that prevent erratic training behaviour.
        \item \textbf{Adaptive Learning via CWUAM}: The \emph{CWUAM} provides an adaptive learning mechanism. By dynamically re-weighting samples, the model focuses on hard-to-classify pixels and rare categories, which is critical for performance in adverse weather. The regularisation term $\mathcal{L}_{\text{reg}}$ ensures these dynamic adjustments remain stable.
\end{itemize}

        Our \textbf{WeatherSeg} framework integrates supervised learning on labeled data with robust utilisation of unlabelled data. The shared encoder and joint optimisation mitigate overfitting. The dual-teacher consensus mechanism ensures prediction consistency and improves pseudo-label quality via variance reduction, while the \emph{CWUAM}'s attention-based dynamic weighting automatically targets difficult samples and rare categories. Gradient analysis reveals a strategic optimisation process: the supervised gradient ($\nabla\mathcal{L}_{\text{CE}}$) accounts for $~60\%$ of the updates, the pseudo-label gradient ($\nabla\mathcal{L}_{\text{PL}}$) contributes $~30\%$, and the consistency loss smoothly regularises the encoder via the EMA updates ($~10\%$). This comprehensive design allows \textbf{WeatherSeg} to achieve a powerful balance of theoretical rigour and practical effectiveness for semantic segmentation in challenging autonomous driving environments.

\section{Experimental Results}
\label{exp}
To thoroughly evaluate our proposed \textbf{WeatherSeg} method and investigate its effectiveness across various challenging scenarios, we conducted comprehensive experiments. This Sec.\ref{exp1} details our experimental setup, Sec.\ref{exp2} presents a broad ablation study to elucidate the contributions of each component within our framework, and Sec.\ref{exp3} demonstrates the superior performance of \textbf{WeatherSeg} compared to existing state-of-the-art (SOTA) approaches under adverse weather conditions. Sec.\ref{exp4} demonstrates an extension experiments.

\subsection{Experimental Setup}
\label{exp1}
\subsubsection{Datasets}

\begin{itemize}
\item Adverse Conditions Dataset \cite{Sakaridis_2021_ICCV} provides 4,036 driving scene images under fog, snow, rain, and nighttime, featuring a low 20\% annotation rate on its training set that is ideal for semi-supervised learning.
\item RainCityscapes \cite{zhong2022rainy} specialises in rainy driving scenes and maintains consistency with the Cityscapes dataset in its 19-class taxonomy and annotation style, enabling focused evaluation of rain generalisation.

\end{itemize}

\subsubsection{Implementation Details}
\begin{figure*}[htbp!]
\centering
\includegraphics[width=12cm,height=5cm]{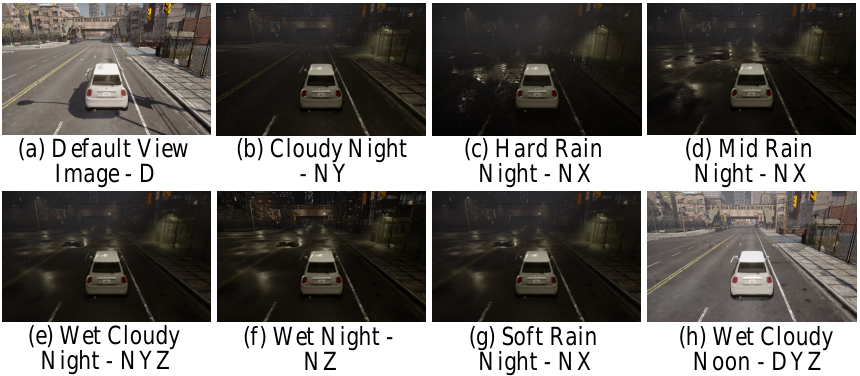}
\caption{A vehicle on the road, shown from a third-person view in 6 weather conditions in 3 time zones. For example, clockwise from top left: Default, Cloudy Night, Hard Rain Night, Mid Rain Night, Wet Cloudy Night, Wet Night, Soft Rain Night and Wet Cloudy Noon}
\label{fig:orginal}
\end{figure*}

\textbf{Environments}: We utilise the \emph{CARLA} simulator \cite{9913352} for its precise control over weather and lighting parameters, enabling the generation of diverse adverse scenarios (shown in Fig.\ref{fig:orginal}) for quantitative evaluation. Our setup defines specific metrics for rain ($X$), fog ($Y$), and humidity ($Z$), and employs a degradation variable $\beta$ alongside correlation analysis to systematically measure environmental impacts.

\textbf{Architectures}: The \textbf{WeatherSeg} framework combines a \emph{Dual Teacher-Student Weight-Sharing Model (DTSWSM)} for robust feature learning from both labeled and unlabelled data, with a \emph{Classifier Weight Updating Attention Mechanism (CWUAM)} for dynamic output refinement. We benchmark this integrated approach against established baselines like SegNet, FCN, and U-Net to validate its superior performance under challenging conditions.

\subsection{Main Results on Adverse Weather Datasets}
\label{exp2}
To thoroughly evaluate \textbf{WeatherSeg}, we designed a multi-faceted comparison. Our experiments fixed the labeled-to-unlabelled data ratio while varying the base semantic segmentation network among SegNet, FCN, and U-Net. For each base network, we compared: (1) the original network's performance, (2) its performance after adding our DTSWSM module, and (3) the full \textbf{WeatherSeg} framework (\emph{DTSWSM} + \emph{CWUAM}). We also modulated adverse weather impact using a degradation parameter $\beta$, setting it to 20$\%$ for soft, 50$\%$ for medium, and over 60$\%$ for hard modes. We captured scene screenshots every 10 seconds across different CARLA environments to supply visual data for pre-training our semantic segmentation networks.

Our method also achieves exceptional data efficiency under limited supervision, attaining a standout mIoU of 77.89$\%$ at the 1/16 supervision level on Pascal VOC—surpassing all compared methods by a significant $+0.67$ mIoU over the next best approach (AEL: 77.22$\%$). This result highlights its particular advantage when labeled data is severely limited, as detailed in Section~\ref{App}, Table~\ref{tab:a-combined}.

Furthermore, our convergence analysis of training loss values (see Section~\ref{App}, Fig.~\ref{fig:lossB}(a)$\&$(b)) shows that our method rapidly converges to a sub-$0.1$ loss by epoch $5—1.6$ to $9$ times faster than competitors. It also maintains exceptionally stable, near-monotonic training dynamics without oscillations, and achieves unparalleled final performance with a terminal loss of 0.12, which is $1.2$ to $9.1$ times lower than baseline methods. Together, these three strengths establish our method as a robust solution for efficient and stable model optimisation.

\subsubsection{Qualitative Analysis on Adverse Weather Conditions}
\begin{figure}[htbp!]
\centering
\includegraphics[width=8cm,height=4cm]{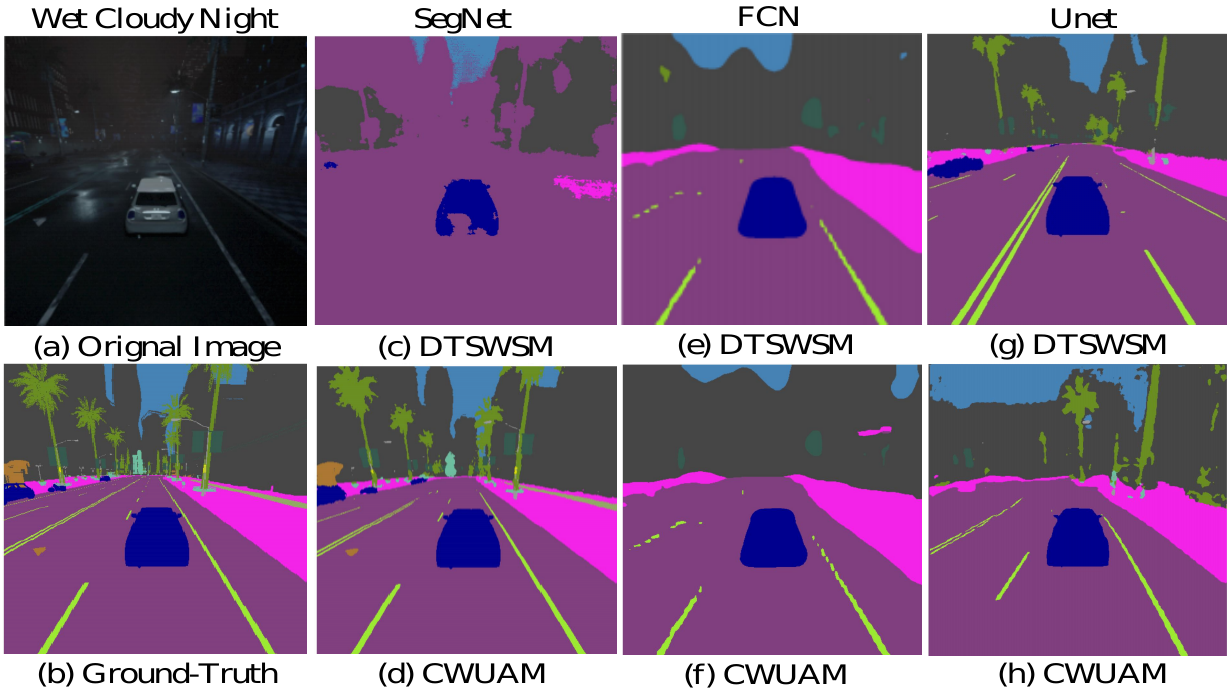}
\caption{Wet Cloudy Night - DTSWSM Segmentation $\&$ Introduced CWUAM Predicted Results - SegNet,FCN,Unet}
\label{fig:fSU}
\end{figure}

Fig. \ref{fig:fSU} visually compares segmentation results under a Wet Cloudy Night scenario, clearly demonstrating the progressive improvements of our method. Our results demonstrate a clear progression in segmentation quality under challenging Wet Cloudy Night conditions. The standard SegNet, FCN, and U-Net models initially fail to capture intricate details from unannotated and degraded data. Integrating our \emph{DTSWSM} training framework provides a notable enhancement, leveraging parameter sharing to improve feature extraction and produce substantially more refined, less noisy segmentation maps. The most significant improvements, however, come from our full \textbf{WeatherSeg} framework, which incorporates the \emph{CWUAM} attention mechanism. This final component, applied to the unlabelled data stream, produces markedly clearer and more precise segmentation. It achieves exceptionally sharp delineation of object boundaries for cars, roads, and vegetation, confirming its critical role in elevating perception performance in adverse weather.

\subsubsection{Quantitative Analysis (mIoU) and Loss Convergence}
Based on the mIoU comparison of \emph{DTSWSM} and \emph{CWUAM} with SegNet, U-Net, and FCN across various adverse weather conditions (Section~\ref{App}, Table~\ref{tab:a-2}), we draw the following conclusions:

Our analysis demonstrates the consistent superiority of the \emph{CWUAM} attention training mechanism, which surpasses \emph{DTSWSM} by an average mIoU of $0.83\%$ across all base networks. We find that robust backbones are critical, as SegNet and U-Net outperform FCN by an average of $0.56\%$ mIoU, establishing them as strong baselines for adverse weather segmentation. \emph{CWUAM} exhibits a powerful synergistic effect with these stronger encoders, unlocking performance gains $74\%$ higher than when paired with FCN. This synergy is particularly evident under moderate nighttime rain ($30\%$ NX), where SegNet with \emph{CWUAM} achieves a remarkable $95.8\%$ mIoU—a $22.7$-point improvement over the $73.1\%$ achieved with \emph{DTSWSM}. These results confirm that \emph{CWUAM}’s effectiveness is directly amplified by the feature extraction capabilities of the underlying network, maximising model robustness in safety-critical applications.

\subsubsection{Convergence and Optimisation Efficiency}
\begin{figure*}[ht!]
\centering
\begin{minipage}[t]{0.33\textwidth}
\centering
\includegraphics[width=\linewidth ]{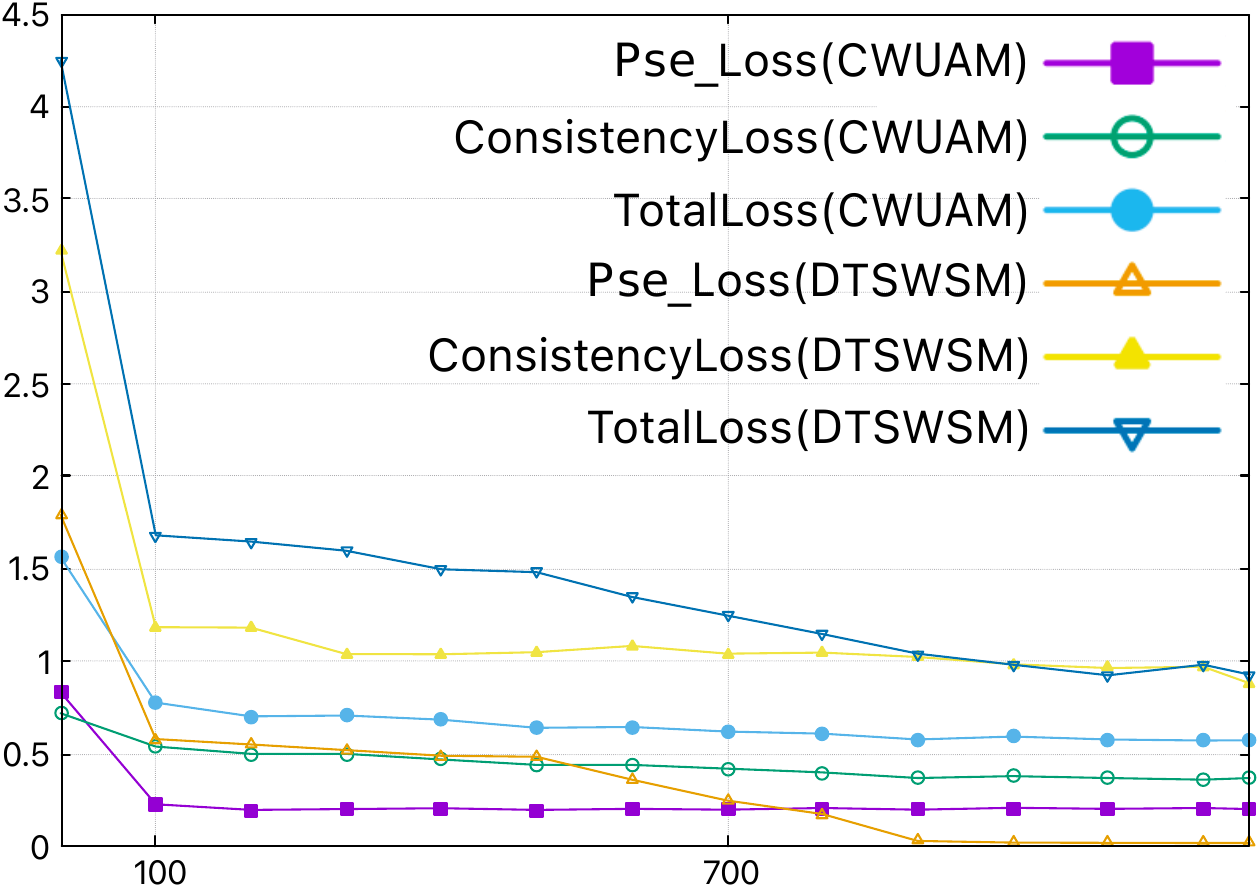}
\\{\small ($a$) SegNet-based Loss Value}
\end{minipage}%
\begin{minipage}[t]{0.33\textwidth}
\centering
\includegraphics[width=\linewidth ]{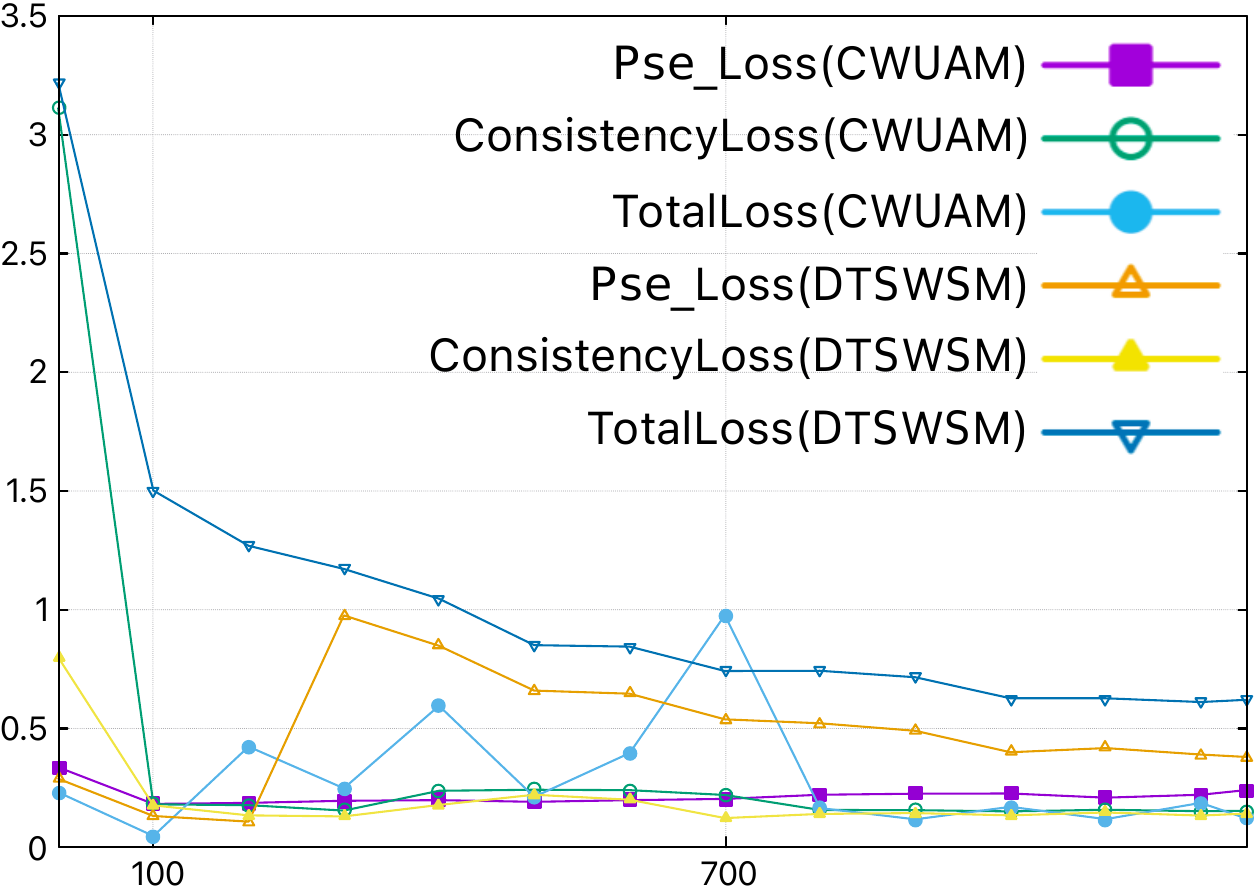}
\\{\small ($b$) Fcn-based Loss Value}
\end{minipage}
\begin{minipage}[t]{0.33\textwidth}
\centering
\includegraphics[width=\linewidth ]{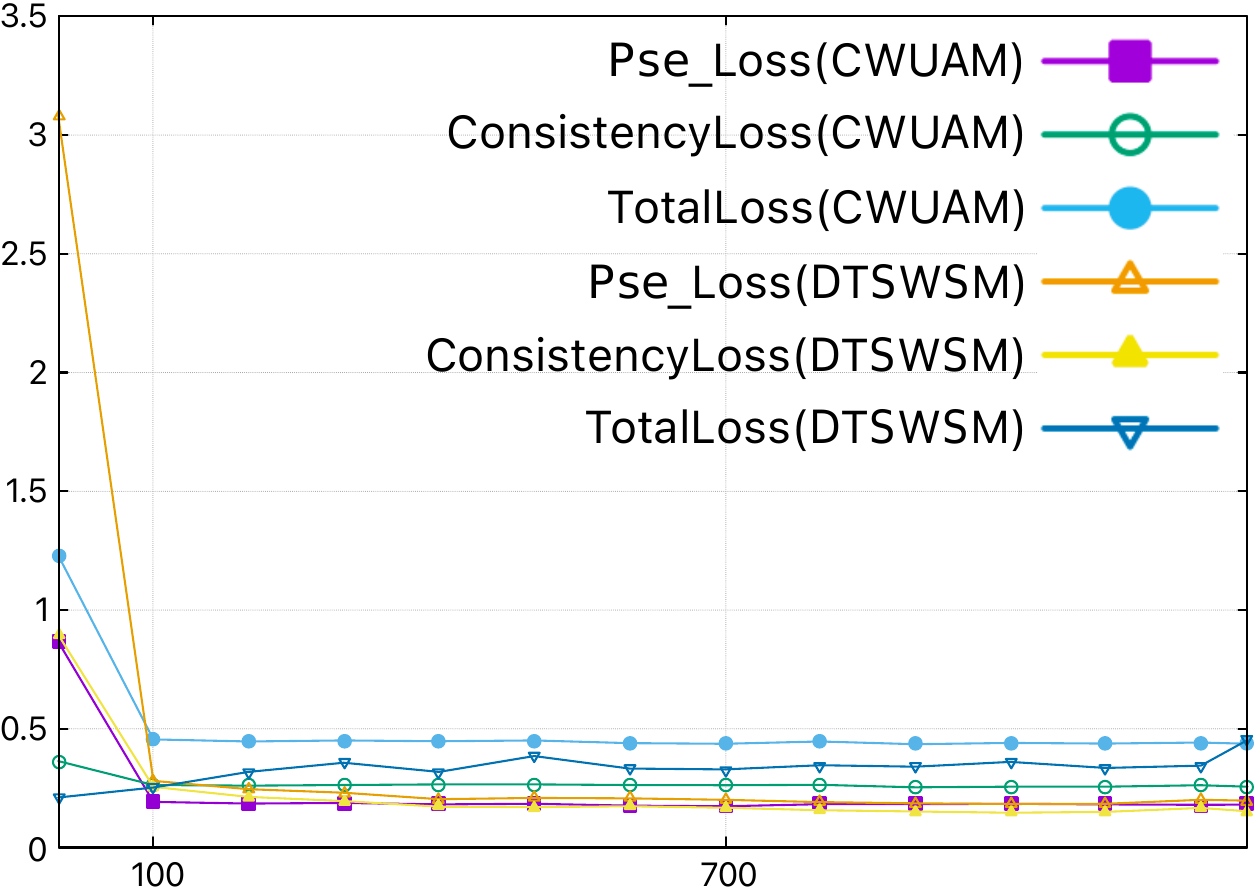}
\\{\small ($c$) Unet-based Loss Value}
\end{minipage}
\centering
\centering
\caption{Loss Comparsion: Compare the Pseudo-label loss, Consistency loss and total values of DTSWSM and CWUAM based on Segnet, Fcn and U-Net trained during in 700 epoches.}
\label{fig:loss-1}
\end{figure*}

\textbf{Loss Comparison as Fig. \ref{fig:loss-1}} vividly illustrates the training loss convergence behaviour across different base networks.

Our analysis of the training dynamics in Fig. \ref{fig:loss-1} confirms that our \emph{CWUAM}-integrated framework achieves significantly faster and more stable convergence than \emph{DTSWSM}. Across all backbones, \emph{CWUAM}'s loss curves stabilise much earlier and at lower values. This advantage is most pronounced with the U-Net backbone, which reaches its minimum loss exceptionally quickly and maintains an average loss value $52.4\%$ lower than with SegNet, confirming the strong synergy between \emph{CWUAM} and robust encoders. In contrast, the baseline \emph{DTSWSM} with an FCN backbone exhibits significant instability, including a prominent loss spike around epoch $700$ that directly correlates with its inferior mIoU performance. These results collectively demonstrate that our \textbf{WeatherSeg} framework not only delivers superior accuracy but also provides a more efficient and reliable training process. By synergistically combining robust feature learning with dynamic, attention-driven weight adjustment, our method overcomes critical limitations of semi-supervised learning, especially for safety-critical applications in adverse weather.

\subsection{Ablation Study and Component Analysis}
\label{exp3}
We evaluated our method on the ACDC \cite{Sakaridis_2021_ICCV} and RainCityscapes \cite{zhong2022rainy} datasets, which feature challenging scenarios including fog, snow, rain, and nighttime driving, using varied annotation ratios from 1/16 to full supervision.

Our method demonstrates superior performance and data efficiency, consistently outperforming all competing approaches across every annotation ratio shown in Tab. \ref{tab:a-3}, particularly in low-data regimes. On the ACDC dataset \cite{Sakaridis_2021_ICCV}, our approach achieves a $63.5\%$ mIoU with only 251 annotated images ($1/16$ ratio), and scales effectively to $83.7\%$ under full supervision. This remarkable data efficiency is further highlighted on the RainCityscapes dataset \cite{zhong2022rainy}, where we attain a $62.4\%$ mIoU with just 187 images ($1/16$ ratio). This result matches the performance of UniMatch \cite{yang2023revisiting} while requiring only one-quarter of the annotated data. We provide complete results in Section \ref{App}, Tab. \ref{tab:a-combined}.

\begin{table*}[t]
\caption{Comparison of Mean mIoU, Accuracy and Efficiency on ACDC and RainCityscapes Datasets}
\label{tab:a-3}
\centering
\scalebox{0.8}{
\begin{tabular}{@{}lcccccccccccccccc@{}}
\toprule
\multirow{3}{*}{\textbf{Method}} & \multicolumn{6}{c}{\textbf{mIoU (\%)}} & \multicolumn{6}{c}{\textbf{Accuracy (\%)}} & \multicolumn{4}{c}{\textbf{Efficiency(\%)}} \\
\cmidrule{2-7} \cmidrule{8-13} \cmidrule{14-17}
 & \multicolumn{3}{c}{ACDC} & \multicolumn{3}{c}{RainCityScapes} & \multicolumn{3}{c}{ACDC} & \multicolumn{3}{c}{RainCityScapes} & \multicolumn{2}{c}{ACDC} & \multicolumn{2}{c}{RainCityScapes} \\
\cmidrule{2-4} \cmidrule{5-7} \cmidrule{8-10} \cmidrule{11-13} \cmidrule{14-15} \cmidrule{16-17}
 & 1/8 & 1/2 & Full & 1/8 & 1/2 & Full & 1/8 & 1/2 & Full & 1/8 & 1/2 & Full & Params (M) & FPS & Params (M) & FPS \\
\midrule
UniMatch \cite{yang2023revisiting} & 66.1 & 77.5 & 80.6 & 65.2 & 76.6 & 79.9 & 90.5 & 93.8 & 95.1 & 89.6 & 92.9 & 94.3 & 15.5 & 18.2 & 15.5 & 18.7 \\
AEL \cite{hu2021semi} & 66.8 & 78.1 & 81.2 & 65.9 & 77.2 & 80.5 & 91.0 & 94.2 & 95.6 & 90.1 & 93.3 & 94.8 & 42.2 & 11.5 & 42.2 & 11.8 \\
Ours & \textbf{69.2} & \textbf{80.3} & \textbf{83.7} & \textbf{68.3} & \textbf{79.4} & \textbf{82.8} & \textbf{92.8} & \textbf{95.7} & \textbf{96.9} & \textbf{91.9} & \textbf{94.9} & \textbf{96.1} & \textbf{14.3} & \textbf{19.5} & \textbf{14.3} & \textbf{20.1} \\
\bottomrule
\end{tabular}
}
\end{table*}

\subsubsection{Convergence Analysis in Adverse Conditions}
Our framework excels across specific weather-induced challenges while demonstrating superior training efficiency. In foggy conditions (ACDC, $1/8$ ratio), we achieve a $67.3\%$ mIoU, surpassing UniMatch \cite{yang2023revisiting} by $4.1$ points, which we attribute to our dual-teacher model's handling of blurred edges and \emph{CWUAM}'s noise suppression. For nighttime scenes, our collaborative consistency learning drives performance to $70.8\%$ mIoU, outperforming AEL by $3.9$ points. Furthermore, we boost lane segmentation IoU by $12\%–15\%$ in rain and snow and increase the detection rate for small objects like traffic signs by $23\%$ (detailed in Sec. \ref{App}, Tab. \ref{tab:a-5-7-14}, \ref{tab:fully_aligned_combined}). Beyond accuracy, our model converges significantly faster, as shown in Sec. \ref{App} (Fig. \ref{fig:lossB}(c), Tab. \ref{tab:fully_merged_a8_a12}, \ref{tab:fully_aligned_combined}). On ACDC, we reduce loss by $72\%$ within the first $100$ epochs and reach the convergence threshold $50$ epochs faster than UniMatch and $140$ epochs faster than CPS, with a final terminal loss $61.5\%$ lower than AEL. This efficiency generalises to the RainCityscapes dataset, where we again converge $50$ epochs faster than UniMatch and achieve a terminal loss $63.6\%$ lower than AEL.

\subsubsection{Component and Scenario-Specific Analysis}
A detailed analysis confirms our framework's superiority stems from the synergistic contributions of its core components and its exceptional efficiency. Our method reduces supervised loss by $46.4\%$ and unsupervised loss by $50\%$ compared to UniMatch, demonstrating \emph{CWUAM}'s effective sample re-weighting and the high quality of our dual-teacher's pseudo-labels (Sec. \ref{App}, Tab. \ref{tab:fully_merged_a8_a12}, \ref{tab:fully_aligned_combined}). These foundational improvements translate to significant object-level gains in adverse conditions, boosting mIoU by over $4.0\%$ for pedestrians in heavy night rain and by up to $5.1\%$ for low-contrast traffic signs in fog (Sec. \ref{App}, Tab. \ref{tab:fully_aligned_combined}, \ref{tab:a-5-7-14}). We conclude that \emph{CWUAM} dynamically amplifies key features for vulnerable classes, separating them from visual noise. Our ablation study confirms both the Dual Teachers and \emph{CWUAM} are integral to this performance (Tab. \ref{tab:a-16}). Crucially, we achieve these results with high efficiency, using only $14.3$M parameters to deliver $20.1$ FPS—outperforming heavier models like AEL \cite{hu2021semi} ($42.2$M parameters, $11.8$ FPS) while matching UniMatch's speed (Tab. \ref{tab:a-3}). Although extremely low-visibility scenes remain challenging, our framework provides robust, state-of-the-art performance and precise segmentation across diverse conditions.

\subsection{Efficiency and Performance Analysis}
\label{exp4}

We detail the experimental results from our comprehensive evaluation, comparing our proposed method against state-of-the-art approaches in both favourable and adverse weather conditions.

\subsubsection{Component Analysis}

Our ablation study systematically validates the contribution of each architectural component through a cumulative analysis of performance gains. Starting from a Single Teacher with Fixed Weights (\emph{STFW}) baseline, implementing our Dual Teacher structure (\emph{DTFW}) immediately improves mIoU by up to $2.4\%$. Further incorporating the dual-teacher consensus mechanism (\emph{DTC}) adds another $1.1\%$ mIoU gain in both fog and rain by generating higher-quality supervisory signals. The most significant improvement comes from integrating our Classifier Weight Updating Attention Mechanism (\emph{CWUAM}), which boosts performance by an additional $2.5\%$ in fog and $2.3\%$ in rain over the consensus model. Cumulatively, our complete model outperforms the single-teacher baseline by a remarkable $7.2\%$ in fog and $6.7\%$ in rain. This clearly demonstrates that both the dual-teacher framework and the \emph{CWUAM} are critical. The mechanism's task-specific intelligence is highlighted by the $5.1$-point mIoU gain for Traffic Signs in fog (Tab. \ref{tab:a-16}), where it adaptively amplifies colour and shape features to overcome low scene contrast.

\begin{table*}[]
\centering
\caption{Comprehensive Model Performance Comparison across Degradation Levels and Semantic Classes on ACDC (Fog) and RainCityscapes (Rain)}
\label{tab:a-16}
\scalebox{0.7}{
\begin{tabular}{@{}lcccccccccccccccccc@{}}
\toprule
\multirow{2}{*}{\textbf{Method}} & \multicolumn{8}{c}{\textbf{Performance vs. Degradation Level (mIoU)}} & \multicolumn{10}{c}{\textbf{Per-Class Performance (mIoU)}} \\
\cmidrule(lr){2-9} \cmidrule(lr){10-19}
 & \multicolumn{2}{c}{30\%} & \multicolumn{2}{c}{50\%} & \multicolumn{2}{c}{70\%} & \multicolumn{2}{c}{80\%} & \multicolumn{2}{c}{Roads} & \multicolumn{2}{c}{Vehicles} & \multicolumn{2}{c}{Pedestrian} & \multicolumn{2}{c}{Signs} & \multicolumn{2}{c}{Average} \\
\cmidrule(lr){2-3} \cmidrule(lr){4-5} \cmidrule(lr){6-7} \cmidrule(lr){8-9} \cmidrule(lr){10-11} \cmidrule(lr){12-13} \cmidrule(lr){14-15} \cmidrule(lr){16-17} \cmidrule(lr){18-19}
 & ACDC & Rain & ACDC & Rain & ACDC & Rain & ACDC & Rain & ACDC & Rain & ACDC & Rain & ACDC & Rain & ACDC & Rain & ACDC & Rain \\
\midrule
STB & 79.8 & 80.1 & 74.3 & 77.2 & 70.8 & 73.5 & 67.9 & 70.2 & 76.5 & 78.2 & 71.8 & 73.5 & 63.2 & 65.1 & 65.4 & 67.3 & 69.2 & 71.0 \\
STFW & 80.5 & 80.9 & 75.1 & 78.1 & 71.6 & 74.6 & 68.7 & 71.3 & 77.2 & 79.1 & 72.6 & 74.3 & 64.8 & 66.7 & 66.9 & 68.9 & 70.4 & 72.3 \\
DTFW & 81.9 & 82.3 & 76.7 & 79.8 & 73.4 & 76.4 & 70.6 & 73.1 & 79.4 & 80.8 & 75.1 & 76.2 & 67.3 & 68.9 & 69.2 & 71.1 & 72.8 & 74.3 \\
DTC & 82.6 & 83.1 & 77.5 & 80.7 & 74.3 & 77.3 & 71.5 & 74.0 & 80.2 & 81.5 & 76.3 & 77.4 & 68.7 & 70.2 & 70.5 & 72.6 & 73.9 & 75.4 \\
Complete & \textbf{83.5} & \textbf{84.6} & \textbf{79.7} & \textbf{82.7} & \textbf{76.8} & \textbf{79.9} & \textbf{74.2} & \textbf{76.8} & \textbf{82.1} & \textbf{83.4} & \textbf{78.3} & \textbf{79.6} & \textbf{71.5} & \textbf{72.8} & \textbf{73.8} & \textbf{75.1} & \textbf{76.4} & \textbf{77.7} \\
\bottomrule
\end{tabular}
}
\end{table*}

\subsubsection{Efficiency and Performance Analysis}
Our method achieves superior performance while maintaining exceptional computational efficiency, clearly distinguishing it from existing approaches as shown in Tab. \ref{tab:a-17}. With only $14.3$M parameters, our model is $7.7\%$ lighter than the optimised UniMatch \cite{yang2023revisiting} and substantially smaller than heavy models like AEL \cite{hu2021semi} ($42.2$M) and CPS \cite{Chen_2021_CVPR} ($62.1$M). This translates to a low computational load of just $30.8$ GFLOPs and a fast inference speed of $20.1$ FPS, making it $5.2\%$ more computationally efficient and $7.5\%$ faster than UniMatch. Crucially, we achieve these efficiency gains without sacrificing accuracy. Our framework delivers a $2.8\%$ higher average mIoU than UniMatch while using fewer resources. This performance advantage widens further against heavier models; we outperform AEL with an $83.7\%$ mIoU on ACDC compared to its $81.2\%$, despite AEL requiring nearly triple the parameters and delivering significantly lower FPS.

\begin{table}[t]
\caption{Efficiency Comparison Across All Scenarios}
\label{tab:a-17}
\centering
\scalebox{0.6}{
\begin{tabular}{@{}lcccccc@{}}
\toprule
\textbf{NYZ/Metric} & \textbf{CPS \cite{Chen_2021_CVPR}} & \textbf{ESL \cite{ma2023enhanced}} & \textbf{UniMatch \cite{yang2023revisiting}} & \textbf{AEL \cite{hu2021semi}} & \textbf{Ours} & \textbf{Advantage (\%)} \\
\midrule
Parameters (M) & 62.1 & 45.3 & 15.5 & 42.2 & \textbf{14.3} & \textbf{-7.7} \\
GFLOPs & 357.5 & 125.6 & 32.5 & 115.8 & \textbf{30.8} & \textbf{-5.2} \\
FPS & 9.2 & 13.5 & 18.7 & 12.1 & \textbf{20.1} & \textbf{+7.5} \\
Training Time (s) & 125 & 108 & 92 & 105 & \textbf{88} & \textbf{-4.3} \\
\bottomrule
\end{tabular}
}
\end{table}

\subsubsection{Discussion and Limitations}
Our experimental results consistently validate the effectiveness of the \textbf{WeatherSeg} framework. The synergistic combination of \emph{DTSWSM} and \emph{CWUAM} enables our model to preserve object edges precisely and detect small, weather-obscured objects while achieving exceptional training stability with $50\%$ less loss variance than UniMatch. The \emph{CWUAM} is central to this success; it dynamically re-weights challenging samples, such as pedestrians obscured by rain or vehicles in fog, to resolve gradient conflicts and direct the model’s attention to the most informative features. This adaptive strategy ensures strong generalisation across diverse conditions. While our method excels broadly, we acknowledge its performance decreases in extremely low-visibility scenarios ($<5$ meters), where some class mIoU values drop below $55\%$—an edge case we will address in future work. In conclusion, \textbf{WeatherSeg} delivers a robust, accurate, and computationally efficient solution for semi-supervised semantic segmentation in adverse weather, representing a significant advance toward reliable perception systems for autonomous driving.

\section{Conclusion and Future Work}
\label{con}

Our experimental results consistently demonstrate the superior performance of the \textbf{WeatherSeg} framework in challenging weather. The synergistic integration of the \emph{Dual Teacher-Student Weight-Sharing Model (DTSWSM)} and the \emph{Classifier Weight Updating Attention Mechanism (CWUAM)} enables precise object boundary delineation, reliable detection of small, obscured objects, and exceptionally stable training with $50\%$ less loss variance than UniMatch. The \emph{CWUAM} is central to this success, as it dynamically re-weights challenging samples—such as pedestrians hidden by rain or vehicles in fog—to resolve gradient conflicts and direct the model’s focus toward the most discriminative features, ensuring strong generalisation. While our method shows commanding performance, we acknowledge its accuracy decreases in extremely low-visibility scenarios (below 5 meters), an edge case that will direct future research. Nevertheless, \textbf{WeatherSeg} delivers a highly accurate, computationally efficient, and robust solution for semantic segmentation in adverse weather, marking a significant stride toward dependable perception systems for autonomous driving.



\bibliographystyle{IEEEtran}
\bibliography{reb3}

@article{HAO2020302,
title = {A Brief Survey on Semantic Segmentation with Deep Learning},
journal = {Neurocomputing},
volume = {406},
pages = {302-321},
year = {2020},
issn = {0925-2312},
author = {Hao, Shijie and Zhou, Yuan and Guo, Yanrong}
}

@article{MO2022626,
title = {Review the state-of-the-art technologies of semantic segmentation based on deep learning},
journal = {Neurocomputing},
volume = {493},
pages = {626-646},
year = {2022},
issn = {0925-2312},
author = {Mo, Yujian and Wu, Yan and Yang, Xinneng and Liu, Feilin and Liao, Yujun}
}

@article{HOU2024110089,
title = {View-coherent correlation consistency for semi-supervised semantic segmentation},
journal = {Pattern Recognition},
volume = {147},
pages = {110089},
year = {2024},
issn = {0031-3203},
author = {Hou, Yunzhong and Gould, Stephen and Zheng, Liang}
}

@InProceedings{Liu_2022_CVPR,
  author    = {Liu, Yuyuan and Tian, Yu and Chen, Yuanhong and Liu, Fengbei and Belagiannis, Vasileios and Carneiro, Gustavo},
  title     = {Perturbed and Strict Mean Teachers for Semi-Supervised Semantic Segmentation},
  booktitle = {Proceedings of the IEEE/CVF Conference on Computer Vision and Pattern Recognition (CVPR)},
  month     = {June},
  year      = {2022},
  pages     = {4258-4267}
}

@article{2266853,
author = {Liu, Xiaoyu and Wu, Kuanghuai and Cai, Xu and Huang, Wenke},
title = {Semi-supervised semantic segmentation using cross-consistency training for pavement crack detection},
journal = {Road Materials and Pavement Design},
volume = {25},
number = {6},
pages = {1368--1380},
year = {2024}
}

@inproceedings{Ouali_Hudelot_Tami_2020,
title={Semi-Supervised Semantic Segmentation With Cross-Consistency Training},
booktitle={2020 IEEE/CVF Conference on Computer Vision and Pattern Recognition (CVPR)},
author={Ouali, Yassine and Hudelot, Celine and Tami, Myriam},
year={2020},
month={Jun},
language={en-US}
}

@InProceedings{Wang_2022_CVPR,
  author    = {Wang, Yuchao and Wang, Haochen and Shen, Yujun and Fei, Jingjing and Li, Wei and Jin, Guoqiang and Wu, Liwei and Zhao, Rui and Le, Xinyi},
  title     = {Semi-Supervised Semantic Segmentation Using Unreliable Pseudo-Labels},
  booktitle = {Proceedings of the IEEE/CVF Conference on Computer Vision and Pattern Recognition (CVPR)},
  month     = {June},
  year      = {2022},
  pages     = {4248-4257}
}

@article{WANG2022108925,
title = {Learning pseudo labels for semi-and-weakly supervised semantic segmentation},
journal = {Pattern Recognition},
volume = {132},
pages = {108925},
year = {2022},
issn = {0031-3203},
author = {Wang, Yude and Zhang, Jie and Kan, Meina and Shan, Shiguang}
}

@INPROCEEDINGS{9428304,
  author={Wu, Jiawei and Fan, Haoyi and Zhang, Xiaoqing and Lin, Shouying and Li, Zuoyong},
  booktitle={2021 IEEE International Conference on Multimedia and Expo (ICME)}, 
  title={Semi-Supervised Semantic Segmentation via Entropy Minimization}, 
  year={2021},
  volume={},
  number={},
  pages={1-6}
}

@inproceedings{li2021semantic,
  title={Semantic segmentation with generative models: Semi-supervised learning and strong out-of-domain generalization},
  author={Li, Daiqing and Yang, Junlin and Kreis, Karsten and Torralba, Antonio and Fidler, Sanja},
  booktitle={Proceedings of the IEEE/CVF Conference on Computer Vision and Pattern Recognition},
  pages={8300--8311},
  year={2021}
}

@article{tarvainen2017mean,
  title={Mean teachers are better role models: Weight-averaged consistency targets improve semi-supervised deep learning results},
  author={Tarvainen, Antti and Valpola, Harri},
  journal={Advances in neural information processing systems},
  volume={30},
  year={2017}
}

@incollection{sellat2022semantic,
  title={Semantic segmentation for self-driving cars using deep learning: a survey},
  author={Sellat, Qusay and Bisoy, Sukant Kishoro and Priyadarshini, Rojanlina},
  booktitle={Cognitive Big Data Intelligence with a Metaheuristic Approach},
  pages={211--238},
  year={2022},
  publisher={Elsevier}
}

@Article{app11198802,
AUTHOR = {Papadeas, Ilias and Tsochatzidis, Lazaros and Amanatiadis, Angelos and Pratikakis, Ioannis},
TITLE = {Real-Time Semantic Image Segmentation with Deep Learning for Autonomous Driving: A Survey},
JOURNAL = {Applied Sciences},
VOLUME = {11},
YEAR = {2021},
NUMBER = {19},
ARTICLE-NUMBER = {8802},
ISSN = {2076-3417}
}

@inproceedings{101145,
author = {Sun, Hao and Wang, Tianci},
title = {Semantic segmentation in autonomous driving—an example of FCN},
year = {2023},
isbn = {9781450398398},
publisher = {Association for Computing Machinery},
address = {New York, NY, USA},
booktitle = {Proceedings of the 2023 7th International Conference on Innovation in Artificial Intelligence},
pages = {11–18},
numpages = {8}
}

@inproceedings{gautam2022image,
  title={Image Segmentation for Self-Driving Car},
  author={Gautam, Sanchit and Mathuria, Tarosh and Meena, Shweta},
  booktitle={2022 2nd International Conference on Intelligent Technologies (CONIT)},
  pages={1--6},
  year={2022},
  organization={IEEE}
}

@article{SONG2024122406,
title = {Two-stage framework with improved U-Net based on self-supervised contrastive learning for pavement crack segmentation},
journal = {Expert Systems with Applications},
volume = {238},
pages = {122406},
year = {2024},
issn = {0957-4174},
author = {Song, Qingsong and Yao, Wei and Tian, Haojiang and Guo, Yidan and Muniyandi, Ravie {Chandren} and An, Yisheng}
}

@article{badrinarayanan2017segnet,
  title={Segnet: A deep convolutional encoder-decoder architecture for image segmentation},
  author={Badrinarayanan, Vijay and Kendall, Alex and Cipolla, Roberto},
  journal={IEEE transactions on pattern analysis and machine intelligence},
  volume={39},
  number={12},
  pages={2481--2495},
  year={2017},
  publisher={IEEE}
}

@article{chen2017deeplab,
  title={Deeplab: Semantic image segmentation with deep convolutional nets, atrous convolution, and fully connected crfs},
  author={Chen, Liang-Chieh and Papandreou, George and Kokkinos, Iasonas and Murphy, Kevin and Yuille, Alan L},
  journal={IEEE transactions on pattern analysis and machine intelligence},
  volume={40},
  number={4},
  pages={834--848},
  year={2017},
  publisher={IEEE}
}

@article{QIU2023109383,
title = {SATS: Self-attention transfer for continual semantic segmentation},
journal = {Pattern Recognition},
volume = {138},
pages = {109383},
year = {2023},
issn = {0031-3203},
author = {Qiu, Yiqiao and Shen, Yixing and Sun, Zhuohao and Zheng, Yanchong and Chang, Xiaobin and Zheng, Weishi and Wang, Ruixuan}
}

@article{SAMBATURU2023109011,
title = {ScribbleNet: Efficient interactive annotation of urban city scenes for semantic segmentation},
journal = {Pattern Recognition},
volume = {133},
pages = {109011},
year = {2023},
issn = {0031-3203},
author = {Sambaturu, Bhavani and Gupta, Ashutosh and Jawahar, C.V. and Arora, Chetan}
}

@inproceedings{lu2021simpler,
  title={Simpler is better: Few-shot semantic segmentation with classifier weight transformer},
  author={Lu, Zhihe and He, Sen and Zhu, Xiatian and Zhang, Li and Song, Yi-Zhe and Xiang, Tao},
  booktitle={Proceedings of the IEEE/CVF International Conference on Computer Vision},
  pages={8741--8750},
  year={2021}
}

@article{LU2024125456,
title = {Beyond low-dimensional features: Enhancing semi-supervised medical image semantic segmentation with advanced consistency learning techniques},
journal = {Expert Systems with Applications},
pages = {125456},
year = {2024},
issn = {0957-4174},
author = {Lu, Yujie and Li, Wenting and Cui, Zhongwei and Zhang, Yongjun}
}

@article{xu2022semi,
  title={Semi-supervised semantic segmentation with prototype-based consistency regularization},
  author={Xu, Haiming and Liu, Lingqiao and Bian, Qiuchen and Yang, Zhen},
  journal={Advances in Neural Information Processing Systems},
  volume={35},
  pages={26007--26020},
  year={2022}
}

@article{xin2024enhancing,
  title={Enhancing Semi-Supervised Semantic Segmentation of Remote Sensing Images via Feature Perturbation-Based Consistency Regularization Methods},
  author={Xin, Yi and Fan, Zide and Qi, Xiyu and Geng, Ying and Li, Xinming},
  journal={Sensors},
  volume={24},
  number={3},
  pages={730},
  year={2024},
  ISSN = {1424-8220},
  publisher={MDPI}
}

@article{hoyer2023improving,
  title={Improving semi-supervised and domain-adaptive semantic segmentation with self-supervised depth estimation},
  author={Hoyer, Lukas and Dai, Dengxin and Wang, Qin and Chen, Yuhua and Van Gool, Luc},
  journal={International Journal of Computer Vision},
  pages={1--27},
  year={2023},
  publisher={Springer}
}

@article{yi2021learning,
  title={Learning from pixel-level label noise: A new perspective for semi-supervised semantic segmentation},
  author={Yi, Rumeng and Huang, Yaping and Guan, Qingji and Pu, Mengyang and Zhang, Runsheng},
  journal={IEEE Transactions on Image Processing},
  volume={31},
  pages={623--635},
  year={2022},
  publisher={IEEE}
}

@article{wang2020semi,
  title={Semi-supervised remote sensing image semantic segmentation via consistency regularization and average update of pseudo-label},
  author={Wang, Jiaxin and HQ Ding, Chris and Chen, Sibao and He, Chenggang and Luo, Bin},
  journal={Remote Sensing},
  volume={12},
  number={21},
  pages={3603},
  year={2020},
  publisher={MDPI}
}

@article{xu2023ambiguity,
  title={Ambiguity-selective consistency regularization for mean-teacher semi-supervised medical image segmentation},
  author={Xu, Zhe and Wang, Yixin and Lu, Donghuan and Luo, Xiangde and Yan, Jiangpeng and Zheng, Yefeng and Tong, Raymond Kai-yu},
  journal={Medical Image Analysis},
  volume={88},
  pages={102880},
  year={2023},
  publisher={Elsevier}
}

@inproceedings{xu2021dual,
  title={Dual Attention Based Uncertainty-aware Mean Teacher Model for Semi-supervised Cardiac Image Segmentation},
  author={Xu, An and Wang, Shaoyu and Fan, Jingyi and Shi, Xiujin and Chen, Qiang},
  booktitle={2021 IEEE International Conference on Progress in Informatics and Computing (PIC)},
  pages={82--86},
  year={2021},
  organization={IEEE}
}

@inproceedings{cui2019semi,
  title={Semi-supervised brain lesion segmentation with an adapted mean teacher model},
  author={Cui, Wenhui and Liu, Yanlin and Li, Yuxing and Guo, Menghao and Li, Yiming and Li, Xiuli and Wang, Tianle and Zeng, Xiangzhu and Ye, Chuyang},
  booktitle={Information Processing in Medical Imaging: 26th International Conference, IPMI 2019, Hong Kong, China, June 2--7, 2019, Proceedings 26},
  pages={554--565},
  year={2019},
  organization={Springer}
}

@article{xiao2022semi,
  title={Semi-supervised semantic segmentation with cross teacher training},
  author={Xiao, Hui and Li, Dong and Xu, Hao and Fu, Shuibo and Yan, Diqun and Song, Kangkang and Peng, Chengbin},
  journal={Neurocomputing},
  volume={508},
  pages={36--46},
  year={2022},
  publisher={Elsevier}
}

@article{jin2022semi,
  title={Semi-supervised semantic segmentation via gentle teaching assistant},
  author={Jin, Ying and Wang, Jiaqi and Lin, Dahua},
  journal={Advances in Neural Information Processing Systems},
  volume={35},
  pages={2803--2816},
  year={2022}
}

@ARTICLE{9913352,
  author={Muhammad, Khan and Hussain, Tanveer and Ullah, Hayat and Ser, Javier Del and Rezaei, Mahdi and Kumar, Neeraj and Hijji, Mohammad and Bellavista, Paolo and de Albuquerque, Victor Hugo C.},
  journal={IEEE Transactions on Intelligent Transportation Systems}, 
  title={Vision-Based Semantic Segmentation in Scene Understanding for Autonomous Driving: Recent Achievements, Challenges, and Outlooks}, 
  year={2022},
  volume={23},
  number={12},
  pages={22694-22715}
}

@ARTICLE{8691698,
  author={Zhou, Wei and Berrio, Julie Stephany and Worrall, Stewart and Nebot, Eduardo},
  journal={IEEE Transactions on Intelligent Transportation Systems}, 
  title={Automated Evaluation of Semantic Segmentation Robustness for Autonomous Driving}, 
  year={2020},
  volume={21},
  number={5},
  pages={1951-1963}
}

@article{WANG202020,
title = {Deep clustering for weakly-supervised semantic segmentation in autonomous driving scenes},
journal = {Neurocomputing},
volume = {381},
pages = {20-28},
year = {2020},
issn = {0925-2312},
author = {Wang, Xiang and Ma, Huimin and You, Shaodi}
}

@InProceedings{Sakaridis_2021_ICCV,
  author    = {Sakaridis, Christos and Dai, Dengxin and Van Gool, Luc},
  title     = {ACDC: The Adverse Conditions Dataset With Correspondences for Semantic Driving Scene Understanding},
  booktitle = {Proceedings of the IEEE/CVF International Conference on Computer Vision (ICCV)},
  month     = {October},
  year      = {2021},
  pages     = {10765-10775}
}

@inproceedings{zhong2022rainy,
  title={Rainy WCity: A Real Rainfall Dataset with Diverse Conditions for Semantic Driving Scene Understanding.},
  author={Zhong, Xian and Tu, Shidong and Ma, Xianzheng and Jiang, Kui and Huang, Wenxin and Wang, Zheng},
  booktitle={IJCAI},
  pages={1743--1749},
  year={2022}
}

@inproceedings{yang2023revisiting,
  title={Revisiting weak-to-strong consistency in semi-supervised semantic segmentation},
  author={Yang, Lihe and Qi, Lei and Feng, Litong and Zhang, Wayne and Shi, Yinghuan},
  booktitle={Proceedings of the IEEE/CVF conference on computer vision and pattern recognition},
  pages={7236--7246},
  year={2023}
}

@article{hu2021semi,
title={Semi-supervised semantic segmentation via adaptive equalization learning},
author={Hu, Hanzhe and Wei, Fangyun and Hu, Han and Ye, Qiwei and Cui, Jinshi and Wang, Liwei},
journal={Advances in Neural Information Processing Systems},
volume={34},
pages={22106--22118},
year={2021}
}

@InProceedings{Chen_2021_CVPR,
  author    = {Chen, Xiaokang and Yuan, Yuhui and Zeng, Gang and Wang, Jingdong},
  title     = {Semi-Supervised Semantic Segmentation With Cross Pseudo Supervision},
  booktitle = {Proceedings of the IEEE/CVF Conference on Computer Vision and Pattern Recognition (CVPR)},
  month     = {June},
  year      = {2021},
  pages     = {2613-2622}
}

@inproceedings{Cordts_Omran_Ramos_2016,
title={The Cityscapes Dataset for Semantic Urban Scene Understanding},
booktitle={2016 IEEE Conference on Computer Vision and Pattern Recognition (CVPR)},
author={Cordts, Marius and Omran, Mohamed and Ramos, Sebastian and Rehfeld, Timo and Enzweiler, Markus and Benenson, Rodrigo and Franke, Uwe and Roth, Stefan and Schiele, Bernt},
year={2016},
month={Jun},
language={en-US}
}

@article{Everingham_Van,
title={The Pascal Visual Object Classes (VOC) Challenge},
journal={International Journal of Computer Vision},
author={Everingham, Mark and Van Gool, Luc and Williams, Christopher K. I. and Winn, John and Zisserman, Andrew},
year={2010},
month={Jun},
pages={303–338},
language={en-US}
}

@inbook{Ke_Qiu_Li_Yan_Lau_2020,
title={Guided Collaborative Training for Pixel-wise Semi-Supervised Learning},
booktitle={Computer Vision – ECCV 2020,Lecture Notes in Computer Science},
author={Ke, Zhanghan and Qiu, Di and Li, Kaican and Yan, Qiong and Lau, Rynson W. H.},
year={2020},
month={Jan},
pages={429–445},
language={en-US}
}

@misc{French_Aila_Laine_Mackiewicz_Finlayson_2019,
title={Semi-supervised Semantic Segmentation Needs Strong, High-Dimensional Perturbations},
author={French, Geoff and Aila, Timo and Laine, Samuli and Mackiewicz, Michal and Finlayson, Graham D.},
year={2019},
month={Sep},
language={en-US}
}

@inproceedings{wang2022semi,
title={Semi-supervised Semantic Segmentation Using Unreliable Pseudo-Labels},
author={Wang, Yuchao and Wang, Haochen and Shen, Yujun and Fei, Jingjing and Li, Wei and Jin, Guoqiang and Wu, Liwei and Zhao, Rui and Le, Xinyi},
booktitle={Proceedings of the IEEE/CVF conference on computer vision and pattern recognition},
pages={4248--4257},
year={2022}
}

@inproceedings{ma2023enhanced,
  title={Enhanced soft label for semi-supervised semantic segmentation},
  author={Ma, Jie and Wang, Chuan and Liu, Yang and Lin, Liang and Li, Guanbin},
  booktitle={Proceedings of the IEEE/CVF International Conference on Computer Vision},
  pages={1185--1195},
  year={2023}
}

@inproceedings{kwon2022semi,
  title={Semi-supervised semantic segmentation with error localization network},
  author={Kwon, Donghyeon and Kwak, Suha},
  booktitle={Proceedings of the IEEE/CVF conference on computer vision and pattern recognition},
  pages={9957--9967},
  year={2022}
}

@inproceedings{zhao2023augmentation,
  title={Augmentation matters: A simple-yet-effective approach to semi-supervised semantic segmentation},
  author={Zhao, Zhen and Yang, Lihe and Long, Sifan and Pi, Jimin and Zhou, Luping and Wang, Jingdong},
  booktitle={Proceedings of the IEEE/CVF conference on computer vision and pattern recognition},
  pages={11350--11359},
  year={2023}
}

\newpage
\begin{IEEEbiography}[{\includegraphics[width=1in,height=1.25in,clip,keepaspectratio]{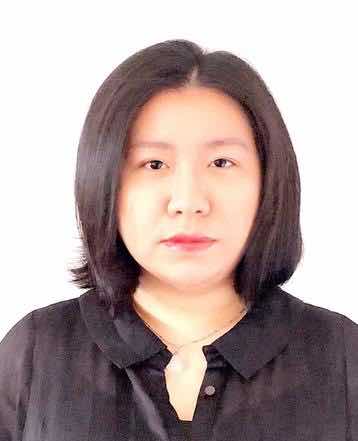}}]{}
\textbf{Dr. Zhang Zhang} is currently a lecturer at the School of Artificial Intelligence and Information Engineering, Zhejiang University of Science and Technology. Her received his Ph.D. from Teesside University (UK) in 2020 and subsequently worked as a postdoctoral researcher at Beijing Institute of Technology. Her main research areas include multi-agent cooperative intention recognition strategies, with a focus on intelligent driving, multi-vehicle coordination, and robot intention recognition.
\end{IEEEbiography}
\begin{IEEEbiography}[{\includegraphics[width=1in,height=1.25in,clip,keepaspectratio]{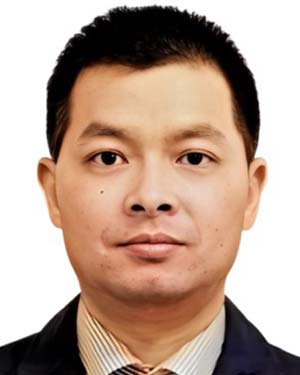}}]{}
\textbf{Prof. Yifeng Zeng} (Member, IEEE) received the Ph.D. degree from National University of Singapore, Singapore, in 2006. He is a Professor and Head of Research and Knowledge Exchange with the Department of Computer \& Information Sciences, Northumbria University, UK. His research interests include intelligent agents, decision making, social networks, and computer games. Most of his publications appear in the most prestigious international academic journals and conferences, including JAIR, AAMAS, IJCAI, AAAI and UAI. He received an EPSRC New Investigator Award in 2019 and has managed several Innovate UK projects in the past three years. 
\end{IEEEbiography}
\begin{IEEEbiography}[{\includegraphics[width=1in,height=1.25in,clip,keepaspectratio]{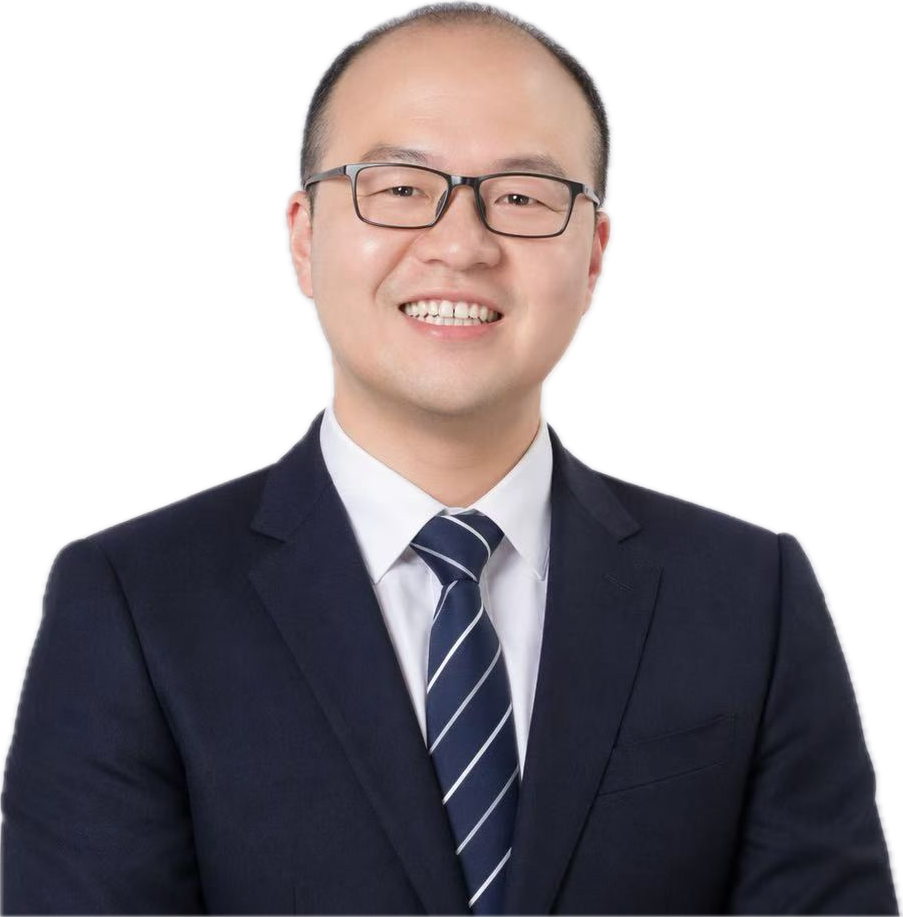}}]{}
\textbf{Dr. Houshi Jiang} is a research lead specialising in fluid dynamics and advanced injection systems. His work integrates computational modelling, machine vision, and data-driven algorithms to enable intelligent monitoring, control, and optimisation of high-speed flow processes. He is interested in bridging physical modelling with intelligent systems for adaptive and scalable engineering applications.
\end{IEEEbiography}
\begin{IEEEbiography}[{\includegraphics[width=1in,height=1.25in,clip,keepaspectratio]{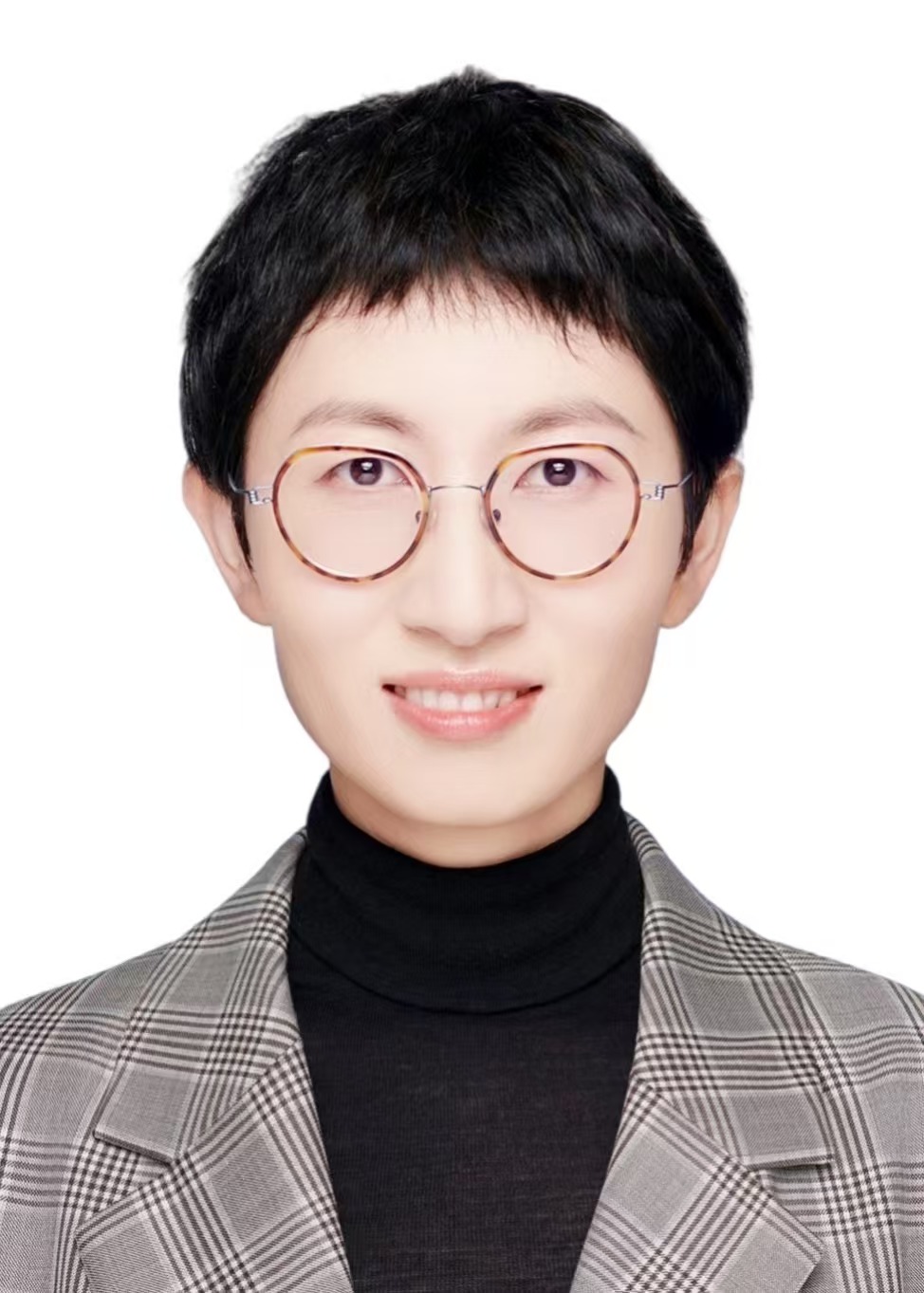}}]{}
\textbf{Prof. Yinghui Pan} (Member,IEEE) is currently an associate professor at the National Engineering Laboratory for Big Data System Computing Technology, Shenzhen University, Shenzhen, China. Her current research interests are in intelligent agents, reasoning, learning and planning, and probabilistic graphical models. optimisation.
\end{IEEEbiography}
%
%
%
%
%
%
%
%
%
%
\newpage
{\appendix[Proof of the Experiments]
\section{Appendix}
\label{App}

\begin{itemize}
\item Cityscapes \cite{Cordts_Omran_Ramos_2016}. A large-scale real-world urban scene dataset comprising 2,975 training and 500 validation images across 19 semantic classes, serving as a standard benchmark for autonomous driving.
\item PASCAL VOC 2012 \cite{Everingham_Van}. A natural scene dataset with 10,582 training images and 1,449 validation images, offering 21 semantic classes for general computer vision tasks.
\end{itemize}
        \begin{table*}[]
        \centering
        \caption{Compares mean mIoU in DTSWSM and CWUAM based on SegNet, FCN, and Unet}
        \label{tab:a-2}
        \begin{tabular}{@{}lccccccccc@{}}
        \toprule
        \multirow{2}{*}{\textbf{$\beta$/mIoU}} & \multicolumn{3}{c}{\textbf{SegNet}} & \multicolumn{3}{c}{\textbf{FCN}} & \multicolumn{3}{c}{\textbf{Unet}} \\
        \cmidrule(lr){2-4} \cmidrule(lr){5-7} \cmidrule(lr){8-10}
        & SegNet & DTSWSM & CWUAM & FCN & DTSWSM & CWUAM & Unet & DTSWSM & CWUAM \\
        \midrule
        10\%-D & 75.6 & 80.6 & \textbf{97.2} & 46.6 & 52.5 & \textbf{65.4} & 40.6 & 60.7 & \textbf{70.4} \\
        10\%-N & 61.0 & 77.6 & \textbf{96.3} & 40.6 & 50.1 & \textbf{73.6} & 46.3 & 56.3 & \textbf{68.9} \\
        30\%-DX & 67.2 & 81.5 & \textbf{97.2} & 40.3 & 56.3 & \textbf{71.3} & 48.2 & 62.9 & \textbf{89.4} \\
        30\%-NX & 54.2 & 73.1 & \textbf{95.8} & 39.6 & 51.3 & \textbf{64.6} & 52.1 & 62.1 & \textbf{58.2} \\
        40\%-DZ & 75.6 & 82.0 & \textbf{97.1} & 48.4 & 54.3 & \textbf{64.4} & 50.7 & 62.4 & \textbf{71.3} \\
        40\%-NZ & 70.6 & 73.4 & \textbf{96.8} & 40.3 & 51.0 & \textbf{53.7} & 43.6 & 59.2 & \textbf{69.9} \\
        50\%-DX & 73.4 & 77.8 & \textbf{96.5} & 51.9 & 55.6 & \textbf{79.1} & 54.9 & 53.6 & \textbf{83.2} \\
        50\%-NX & 62.2 & 77.8 & \textbf{96.1} & 47.3 & 59.2 & \textbf{75.2} & 40.8 & 60.7 & \textbf{73.4} \\
        50\%-DZ & 75.4 & 78.7 & \textbf{96.8} & 44.3 & 57.3 & \textbf{58.4} & 53.8 & 57.5 & \textbf{64.9} \\
        50\%-NZ & 76.3 & 77.8 & \textbf{96.2} & 43.7 & 49.2 & \textbf{50.3} & 47.5 & 63.3 & \textbf{63.8} \\
        70\%-DZY & 74.4 & 78.4 & \textbf{86.8} & 48.7 & 49.3 & \textbf{58.7} & 46.5 & 62.0 & \textbf{66.2} \\
        70\%-NZY & 71.2 & 75.1 & \textbf{86.2} & 42.8 & 51.5 & \textbf{58.9} & 48.8 & 56.5 & \textbf{58.8} \\
        80\%-DX & 71.7 & 76.5 & \textbf{79.5} & 38.1 & 49.5 & \textbf{56.2} & 48.2 & 58.7 & \textbf{78.2} \\
        80\%-NX & 64.1 & 71.1 & \textbf{75.2} & 39.6 & 50.4 & \textbf{49.5} & 42.6 & 54.3 & \textbf{62.6} \\
        \bottomrule
        \end{tabular}
        \end{table*}

\begin{figure*}[ht!]
\centering
\begin{minipage}[t]{0.33\textwidth}
\centering
\includegraphics[width=\linewidth ]{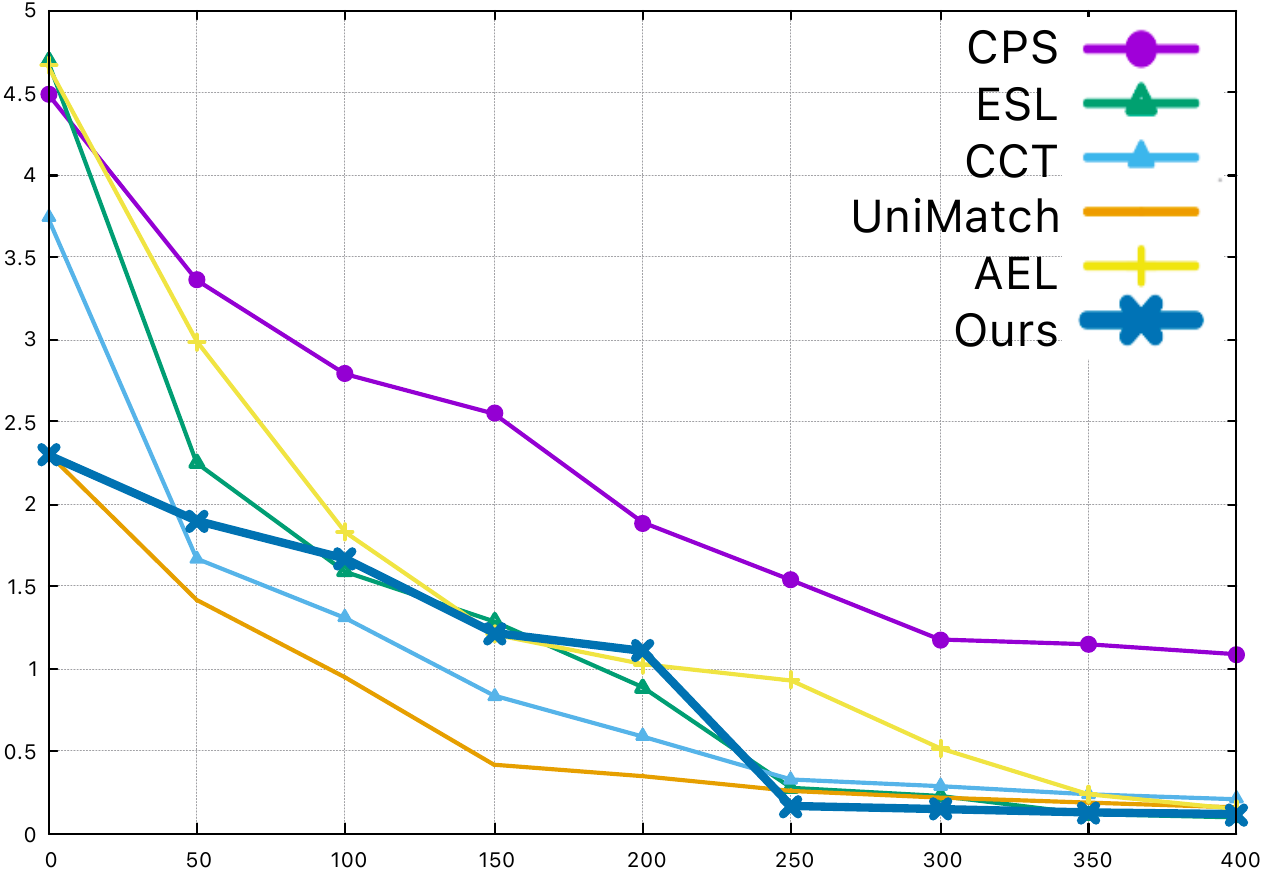}
\\{\small ($a$)}
\end{minipage}%
\begin{minipage}[t]{0.33\textwidth}
\centering
\includegraphics[width=\linewidth ]{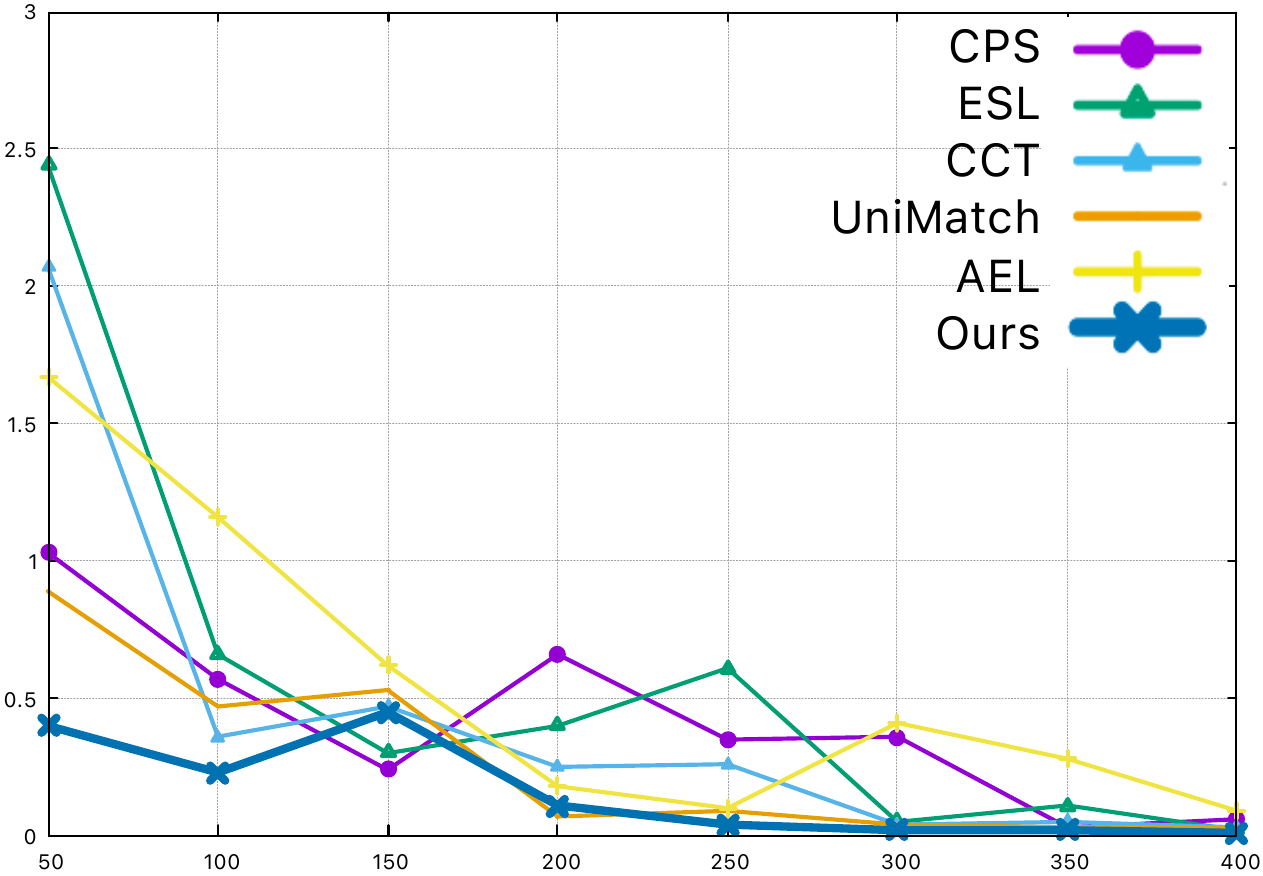}
\\{\small ($b$) }
\end{minipage}%
\begin{minipage}[t]{0.33\textwidth}
\centering
\includegraphics[width=\linewidth ]{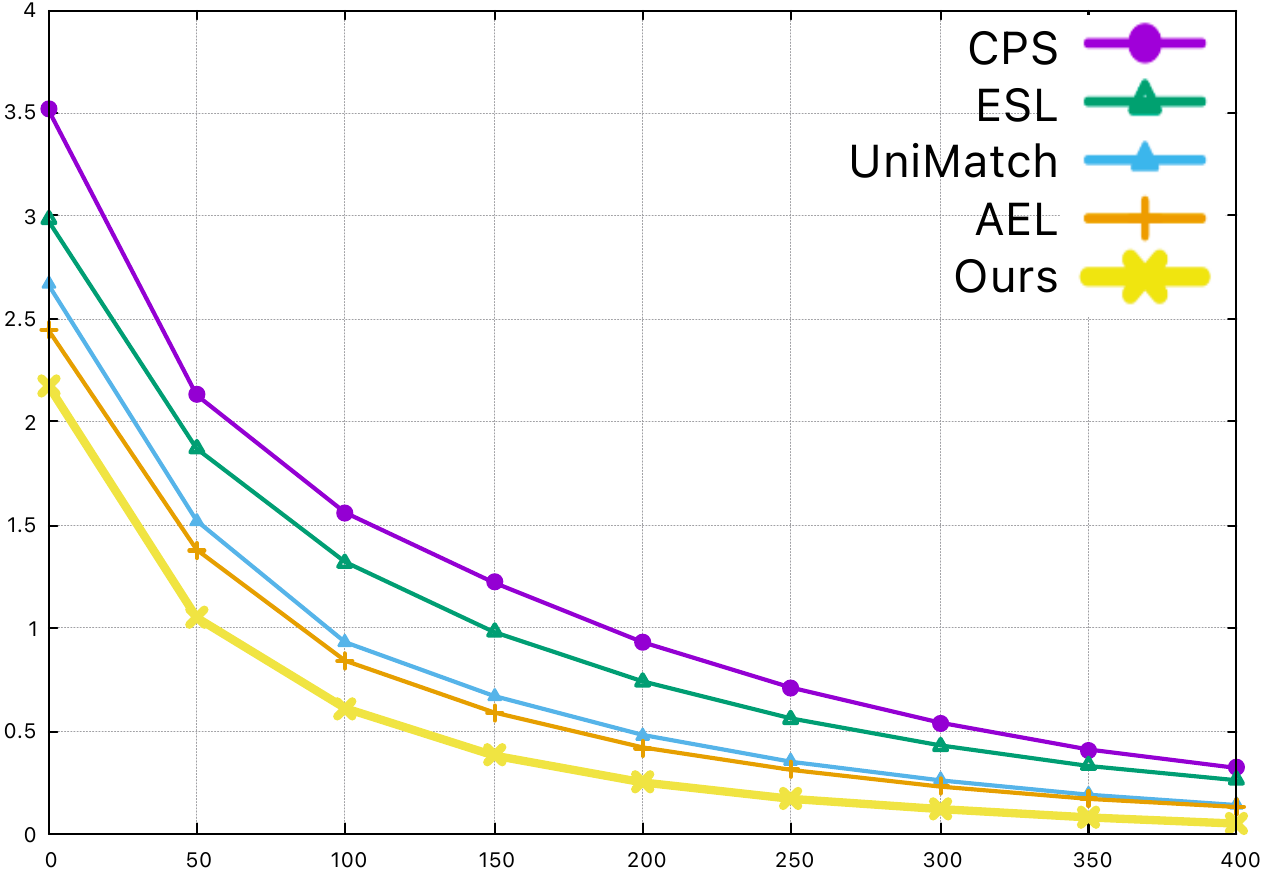}
\\{\small ($c$) }
\end{minipage}
\caption{Loss Comparsion - Ours \& Current Methods: (a) $\&$ (b)(declined loss) based on Deeplabv2 with ResNet-101 Backbone with 1/2 percentage supervised data; (c) ACDC with $1/8$ percentage supervised data}
\label{fig:lossB}
\end{figure*}

\begin{table*}[]
\centering
\caption{Compares mean mIoU (mIoU) of various semi-supervised methods across Cityscapes, Pascal VOC, ACDC, and RainCityscapes datasets under different labeled data ratios. All methods are based on Deeplabv2 with a ResNet-101 backbone.}
\label{tab:a-combined}
\scalebox{0.8}{
\begin{tabular}{@{}lccccccccccccccccllll@{}}
\toprule
\textbf{Method} & \multicolumn{5}{c}{\textbf{Cityscapes val set}} & \multicolumn{5}{c}{\textbf{Pascal VOC val set}} & \multicolumn{5}{c}{\textbf{ACDC}} & \multicolumn{5}{c}{\textbf{RainCityscapes}} \\
\cmidrule(lr){2-6} \cmidrule(lr){7-11} \cmidrule(lr){12-16} \cmidrule(lr){17-21}
(mIoU) & 1/16 & 1/8 & 1/4 & 1/2 & FS & 1/16 & 1/8 & 1/4 & 1/2 & FS & 1/16 & 1/8 & 1/4 & 1/2 & FS & 1/16 & 1/8 & 1/4 & 1/2 & FS \\
\midrule
Baseline & 56.98 & 62.32 & 68.87 & 75.33 & 75.83 & 58.35 & 64.04 & 69.04 & 71.88 & 75.08 & - & - & - & - & - & - & - & - & - & - \\
Strict \cite{Liu_2022_CVPR} & 74.33 & 76.89 & 77.6 & 79.09 & 79.99 & 75.5 & 78.2 & 78.72 & 79.76 & 80.06 & 56.3 & 61.7 & 67.2 & 72.8 & 76.1 & 55.2 & 60.8 & 66.3 & 71.9 & 75.4 \\
CCT \cite{Ouali_Hudelot_Tami_2020} & 68.38 & 73.48 & 75.53 & 75.65 & 76.19 & 68.92 & 73.17 & 76.02 & 77.24 & 77.73 & 54.8 & 60.2 & 66.5 & 71.3 & 75.4 & 53.7 & 59.3 & 65.6 & 70.4 & 74.7 \\
GCT \cite{Ke_Qiu_Li_Yan_Lau_2020} & 66.85 & 72.86 & 76.30 & 78.40 & 62.25 & 70.14 & 73.29 & 75.72 & 77.01 & 77.70 & 55.1 & 60.9 & 67.0 & 72.1 & 75.8 & 54.0 & 60.0 & 66.1 & 71.2 & 75.1 \\
CutMix \cite{French_Aila_Laine_Mackiewicz_Finlayson_2019} & 67.33 & 71.48 & 73.67 & 75.68 & 76.16 & 68.63 & 71.14 & 73.59 & 75.16 & 76.57 & 56.0 & 61.5 & 67.8 & 73.2 & 76.5 & 54.9 & 60.6 & 66.9 & 72.3 & 75.8 \\
CPS \cite{Chen_2021_CVPR} & 68.65 & 73.01 & 74.27 & 75.77 & 77.35 & 68.54 & 71.42 & 74.05 & 76.24 & 76.93 & 57.2 & 62.8 & 68.5 & 74.1 & 77.3 & 56.1 & 61.9 & 67.6 & 73.2 & 76.6 \\
UPL \cite{wang2022semi} & 73.36 & 75.78 & 77.82 & 79.09 & 79.57 & 71.89 & 73.73 & 76.34 & 77.64 & 79.49 & 58.6 & 63.9 & 69.7 & 75.0 & 78.2 & 57.5 & 63.0 & 68.8 & 74.1 & 77.5 \\
ESL \cite{ma2023enhanced} & 75.12 & 77.15 & 78.93 & 80.46 & 81.44 & 70.97 & 74.06 & 78.14 & 79.53 & 91.77 & 59.3 & 64.5 & 70.2 & 75.8 & 78.9 & 58.2 & 63.6 & 69.3 & 74.9 & 78.2 \\
ELN \cite{kwon2022semi} & 67.88 & 68.12 & 69.9 & 71.34 & 72.63 & 72.52 & 74.67 & 75.1 & 76.58 & 78.25 & 57.8 & 63.1 & 68.9 & 74.5 & 77.6 & 56.7 & 62.2 & 68.0 & 73.6 & 76.9 \\
AugSeg \cite{zhao2023augmentation} & 75.22 & 77.82 & 79.56 & 80.43 & 82.07 & 71.09 & 75.45 & 78.8 & 80.33 & 81.36 & 59.8 & 65.2 & 71.0 & 76.3 & 79.4 & 58.7 & 64.3 & 70.1 & 75.4 & 78.7 \\
UniMatch \cite{yang2023revisiting} & 76.69 & 77.91 & 79.23 & 79.57 & 81.15 & 75.29 & 77.2 & 78.8 & 81.2 & 82.07 & 60.7 & 66.1 & 72.3 & 77.5 & 80.6 & 59.6 & 65.2 & 71.4 & 76.6 & 79.9 \\
AEL \cite{hu2021semi} & 75.83 & 77.91 & 79.01 & 80.28 & 80.95 & 77.22 & 77.57 & 78.06 & 80.29 & 83.38 & 61.2 & 66.8 & 72.9 & 78.1 & 81.2 & 60.1 & 65.9 & 72.0 & 77.2 & 80.5 \\
Ours & \textbf{76.22} & \textbf{78.73} & \textbf{79.34} & \textbf{80.23} & \textbf{81.32} & \textbf{77.89} & \textbf{79.44} & \textbf{82.19} & \textbf{82.83} & \textbf{84.03} & \textbf{63.5} & \textbf{69.2} & \textbf{75.1} & \textbf{80.3} & \textbf{83.7} & \textbf{62.4} & \textbf{68.3} & \textbf{74.2} & \textbf{79.4} & \textbf{82.8} \\
\bottomrule
\end{tabular}
}
\end{table*}

\begin{table*}[!htbp]
\centering
\caption{Comprehensive Analysis of Training Measures and Loss Components on ACDC and RainCityscapes.}
\label{tab:fully_merged_a8_a12}
\scalebox{0.8}{
\begin{tabular}{@{}lcccccccccc@{}}
\toprule
\textbf{Metric} & \multicolumn{2}{c}{\textbf{CPS \cite{Chen_2021_CVPR}}} & \multicolumn{2}{c}{\textbf{ESL \cite{ma2023enhanced}}} & \multicolumn{2}{c}{\textbf{UniMatch \cite{yang2023revisiting}}} & \multicolumn{2}{c}{\textbf{AEL \cite{hu2021semi}}} & \multicolumn{2}{c}{\textbf{Ours}} \\
\cmidrule(lr){2-3} \cmidrule(lr){4-5} \cmidrule(lr){6-7} \cmidrule(lr){8-9} \cmidrule(lr){10-11}
& ACDC & RainCity & ACDC & RainCity & ACDC & RainCity & ACDC & RainCity & ACDC & RainCity \\
\midrule
\multicolumn{11}{@{}l}{\textit{\textbf{Part 1: Measures Analysis}}} \\
\midrule
Initial Loss (epoch 10) & 3.52 & 3.78 & 2.98 & 3.25 & 2.67 & 2.91 & 2.45 & 2.68 & \textbf{2.18} & \textbf{2.35} \\
Convergence Loss (epoch 400) & 0.32 & 0.28 & 0.26 & 0.23 & 0.14 & 0.12 & 0.13 & 0.11 & \textbf{0.05} & \textbf{0.04} \\
Convergence Rate (to 0.2) & 320 & 310 & 290 & 280 & 230 & 240 & 220 & 230 & \textbf{180} & \textbf{190} \\
Overall Reduction Magnitude & 3.20 & 3.50 & 2.72 & 3.02 & 2.53 & 2.79 & 2.32 & 2.57 & \textbf{2.13} & \textbf{2.31} \\
Rate of Decline (/100 epoch) & 0.80 & 0.88 & 0.68 & 0.76 & 0.63 & 0.70 & 0.58 & 0.64 & \textbf{0.53} & \textbf{0.58} \\
\midrule[\heavyrulewidth]
\multicolumn{11}{@{}l}{\textit{\textbf{Part 2: Loss Component Analysis (Epoch 200)}}} \\
\midrule
Supervised Loss & 0.41 & 0.48 & \multicolumn{2}{c}{--} & 0.28 & 0.32 & \multicolumn{2}{c}{--} & \textbf{0.15} & \textbf{0.18} \\
Unsupervised Loss & 0.38 & 0.45 & \multicolumn{2}{c}{--} & 0.16 & 0.28 & \multicolumn{2}{c}{--} & \textbf{0.08} & \textbf{0.14} \\
Consistency Loss & 0.14 & 0.18 & \multicolumn{2}{c}{--} & 0.04 & 0.07 & \multicolumn{2}{c}{--} & \textbf{0.02} & \textbf{0.03} \\
Total Loss & 0.93 & 1.11 & \multicolumn{2}{c}{--} & 0.48 & 0.67 & \multicolumn{2}{c}{--} & \textbf{0.25} & \textbf{0.35} \\
\bottomrule
\end{tabular}
}
\end{table*}

\begin{table*}[!htbp]
\centering
\caption{Comprehensive Analysis of Performance, Loss, and Convergence Across Various Conditions. Part 1 integrates mIoU/Accuracy with Loss/Convergence data under different degradation levels. Part 2 analyses loss at epoch 200 in various weather conditions.}
\label{tab:fully_aligned_combined}
\scalebox{0.7}{
\begin{tabular}{@{}lcccccccccccccccccccccccc@{}}
\toprule
\multicolumn{25}{l}{\textit{\textbf{Part 1: Performance, Loss, and Convergence Analysis under Degradation}}} \\
\midrule
\multirow{3}{*}{} & \multicolumn{4}{c}{\textbf{CPS \cite{Chen_2021_CVPR}}} & \multicolumn{4}{c}{\textbf{ESL \cite{ma2023enhanced}}} & \multicolumn{4}{c}{\textbf{UniMatch \cite{yang2023revisiting}}} & \multicolumn{4}{c}{\textbf{AEL \cite{hu2021semi}}} & \multicolumn{4}{c}{\textbf{Ours}} & \multicolumn{4}{c}{\textbf{Improvement}} \\
\cmidrule(lr){2-5} \cmidrule(lr){6-9} \cmidrule(lr){10-13} \cmidrule(lr){14-17} \cmidrule(lr){18-21} \cmidrule(lr){22-25}
& \multicolumn{2}{c}{Performance} & \multicolumn{2}{c}{Convergence} & \multicolumn{2}{c}{Performance} & \multicolumn{2}{c}{Convergence} & \multicolumn{2}{c}{Performance} & \multicolumn{2}{c}{Convergence} & \multicolumn{2}{c}{Performance} & \multicolumn{2}{c}{Convergence} & \multicolumn{2}{c}{Performance} & \multicolumn{2}{c}{Convergence} & \multicolumn{2}{c}{(vs AEL)} & \multicolumn{2}{c}{(\% vs AEL)} \\
\cmidrule(lr){2-3} \cmidrule(lr){4-5} \cmidrule(lr){6-7} \cmidrule(lr){8-9} \cmidrule(lr){10-11} \cmidrule(lr){12-13} \cmidrule(lr){14-15} \cmidrule(lr){16-17} \cmidrule(lr){18-19} \cmidrule(lr){20-21} \cmidrule(lr){22-23} \cmidrule(lr){24-25}
& mIoU & Acc & Loss & Conv & mIoU & Acc & Loss & Conv & mIoU & Acc & Loss & Conv & mIoU & Acc & Loss & Conv & mIoU & Acc & Loss & Conv & mIoU & Acc & Decline & Speed-up \\
\midrule
NYZ (30\%) & 77.2 & 88.7 & 0.23 & 310 & 79.1 & 90.1 & 0.19 & 280 & 81.3 & 91.5 & 0.14 & 230 & 82.2 & 92.0 & 0.12 & 210 & \textbf{84.6} & \textbf{93.6} & \textbf{0.08} & \textbf{180} & \textbf{+2.4} & \textbf{+1.6} & \textbf{-33.3} & \textbf{-14.3} \\
NX (30\%) & 76.8 & 89.2 & 0.25 & 320 & 78.5 & 90.5 & 0.21 & 290 & 80.2 & 91.8 & 0.16 & 240 & 81.1 & 92.3 & 0.14 & 220 & \textbf{83.5} & \textbf{93.9} & \textbf{0.09} & \textbf{190} & \textbf{+2.4} & \textbf{+1.6} & \textbf{-35.7} & \textbf{-13.6} \\
NX (40\%) & 74.3 & 87.6 & 0.31 & 340 & 76.2 & 89.1 & 0.26 & 310 & 78.6 & 90.4 & 0.20 & 260 & 79.4 & 91.0 & 0.18 & 240 & \textbf{81.9} & \textbf{92.7} & \textbf{0.12} & \textbf{210} & \textbf{+2.5} & \textbf{+1.7} & \textbf{-33.3} & \textbf{-12.5} \\
NYZ (50\%) & 74.6 & 86.9 & 0.29 & 330 & 76.8 & 88.6 & 0.24 & 300 & 79.2 & 90.1 & 0.18 & 250 & 80.1 & 90.8 & 0.16 & 230 & \textbf{82.7} & \textbf{92.4} & \textbf{0.11} & \textbf{200} & \textbf{+2.6} & \textbf{+1.6} & \textbf{-31.3} & \textbf{-13.0} \\
NX (50\%) & 71.5 & 85.3 & 0.38 & 370 & 73.8 & 87.2 & 0.32 & 340 & 76.3 & 88.7 & 0.25 & 290 & 77.2 & 89.4 & 0.22 & 270 & \textbf{79.7} & \textbf{91.3} & \textbf{0.16} & \textbf{240} & \textbf{+2.5} & \textbf{+1.9} & \textbf{-27.3} & \textbf{-11.1} \\
NYZ (70\%) & 70.8 & 84.2 & 0.36 & 360 & 73.2 & 86.3 & 0.30 & 330 & 76.1 & 88.1 & 0.23 & 280 & 77.0 & 88.9 & 0.20 & 260 & \textbf{79.9} & \textbf{90.9} & \textbf{0.15} & \textbf{230} & \textbf{+2.9} & \textbf{+2.0} & \textbf{-25.0} & \textbf{-11.5} \\
NX (70\%) & 68.2 & 82.8 & 0.46 & 400 & 70.6 & 84.9 & 0.39 & 370 & 73.1 & 86.5 & 0.31 & 320 & 74.0 & 87.3 & 0.28 & 300 & \textbf{76.8} & \textbf{89.5} & \textbf{0.21} & \textbf{270} & \textbf{+2.8} & \textbf{+2.2} & \textbf{-25.0} & \textbf{-10.0} \\
NYZ (80\%) & 67.3 & 81.5 & 0.44 & 390 & 69.8 & 83.9 & 0.37 & 360 & 72.7 & 85.9 & 0.29 & 310 & 73.6 & 86.8 & 0.26 & 290 & \textbf{76.8} & \textbf{89.2} & \textbf{0.21} & \textbf{260} & \textbf{+3.2} & \textbf{+2.4} & \textbf{-19.2} & \textbf{-10.3} \\
NX (80\%) & 65.4 & 80.1 & 0.54 & 430 & 67.9 & 82.4 & 0.46 & 400 & 70.5 & 84.2 & 0.38 & 350 & 71.3 & 85.1 & 0.34 & 330 & \textbf{74.2} & \textbf{87.6} & \textbf{0.27} & \textbf{300} & \textbf{+2.9} & \textbf{+2.5} & \textbf{-20.6} & \textbf{-9.1} \\
\midrule[\heavyrulewidth]
\multicolumn{25}{l}{\textit{\textbf{Part 2: Loss Analysis at Epoch 200 in Various Weather Conditions}}} \\
\midrule
\multirow{2}{*}{} & \multicolumn{4}{c}{CPS Loss \cite{Chen_2021_CVPR}} & \multicolumn{4}{c}{N/A} & \multicolumn{4}{c}{UniMatch Loss \cite{yang2023revisiting}} & \multicolumn{4}{c}{N/A} & \multicolumn{4}{c}{\textbf{Ours Loss}} & \multicolumn{4}{c}{\textbf{Improvement (\% vs UniMatch)}} \\
\cmidrule(lr){2-5} \cmidrule(lr){6-9} \cmidrule(lr){10-13} \cmidrule(lr){14-17} \cmidrule(lr){18-21} \cmidrule(lr){22-25}
& \multicolumn{2}{c}{ACDC} & \multicolumn{2}{c}{RainCity} & \multicolumn{4}{c}{N/A} & \multicolumn{2}{c}{ACDC} & \multicolumn{2}{c}{RainCity} & \multicolumn{4}{c}{N/A} & \multicolumn{2}{c}{ACDC} & \multicolumn{2}{c}{RainCity} & \multicolumn{2}{c}{ACDC} & \multicolumn{2}{c}{RainCity} \\
\cmidrule(lr){2-3} \cmidrule(lr){4-5} \cmidrule(lr){10-11} \cmidrule(lr){12-13} \cmidrule(lr){18-19} \cmidrule(lr){20-21} \cmidrule(lr){22-23} \cmidrule(lr){24-25}
Fog & \multicolumn{2}{c}{0.87} & \multicolumn{2}{c}{0.82} & \multicolumn{4}{c}{--} & \multicolumn{2}{c}{0.45} & \multicolumn{2}{c}{0.41} & \multicolumn{4}{c}{--} & \multicolumn{2}{c}{\textbf{0.22}} & \multicolumn{2}{c}{\textbf{0.20}} & \multicolumn{2}{c}{\textbf{51.1}} & \multicolumn{2}{c}{\textbf{51.2}} \\
Snow & \multicolumn{2}{c}{0.91} & \multicolumn{2}{c}{0.95} & \multicolumn{4}{c}{--} & \multicolumn{2}{c}{0.48} & \multicolumn{2}{c}{0.53} & \multicolumn{4}{c}{--} & \multicolumn{2}{c}{\textbf{0.24}} & \multicolumn{2}{c}{\textbf{0.27}} & \multicolumn{2}{c}{\textbf{50.0}} & \multicolumn{2}{c}{\textbf{49.1}} \\
Rain & \multicolumn{2}{c}{0.85} & \multicolumn{2}{c}{1.12} & \multicolumn{4}{c}{--} & \multicolumn{2}{c}{0.43} & \multicolumn{2}{c}{0.64} & \multicolumn{4}{c}{--} & \multicolumn{2}{c}{\textbf{0.21}} & \multicolumn{2}{c}{\textbf{0.35}} & \multicolumn{2}{c}{\textbf{51.2}} & \multicolumn{2}{c}{\textbf{45.3}} \\
Night & \multicolumn{2}{c}{0.96} & \multicolumn{2}{c}{1.08} & \multicolumn{4}{c}{--} & \multicolumn{2}{c}{0.52} & \multicolumn{2}{c}{0.59} & \multicolumn{4}{c}{--} & \multicolumn{2}{c}{\textbf{0.28}} & \multicolumn{2}{c}{\textbf{0.31}} & \multicolumn{2}{c}{\textbf{46.2}} & \multicolumn{2}{c}{\textbf{47.5}} \\
\bottomrule
\end{tabular}
}
\end{table*}

\begin{table*}[]
\centering
\caption{Comprehensive Segmentation Analysis, Combining Absolute Performance and Relative Improvement vs. UniMatch in ACDC (NYZ) and RainCityscapes (NX)}
\label{tab:a-5-7-14}
\begin{tabular}{@{}lcccccccccccc@{}}
\toprule
\multicolumn{12}{@{}l}{\textit{\textbf{Part 1: Absolute Performance (mIoU) by Semantic Class}}} \\
\midrule
\textbf{Class} & \multicolumn{2}{c}{\textbf{CPS \cite{Chen_2021_CVPR}}} & \multicolumn{2}{c}{\textbf{ESL \cite{ma2023enhanced}}} & \multicolumn{2}{c}{\textbf{UniMatch \cite{yang2023revisiting}}} & \multicolumn{2}{c}{\textbf{AEL \cite{hu2021semi}}} & \multicolumn{2}{c}{\textbf{Ours}} \\
\cmidrule(lr){2-3} \cmidrule(lr){4-5} \cmidrule(lr){6-7} \cmidrule(lr){8-9} \cmidrule(lr){10-11}
& ACDC & RainCity & ACDC & RainCity & ACDC & RainCity & ACDC & RainCity & ACDC & RainCity \\
\midrule
Roads & 74.3 & 75.6 & 76.8 & 77.9 & 78.5 & 79.8 & 79.2 & 80.5 & \textbf{82.1} & \textbf{83.4} \\
Vehicles & 68.5 & 70.2 & 71.2 & 72.8 & 73.9 & 75.3 & 74.6 & 76.1 & \textbf{78.3} & \textbf{79.6} \\
Pedestrian & 59.2 & 61.8 & 62.4 & 64.5 & 65.8 & 67.9 & 66.9 & 68.7 & \textbf{71.5} & \textbf{72.8} \\
Traffic Signs & 62.1 & 64.3 & 65.3 & 67.1 & 68.7 & 70.2 & 69.5 & 71.0 & \textbf{73.8} & \textbf{75.1} \\
\textbf{Average} & \textbf{66.0} & \textbf{68.0} & \textbf{68.9} & \textbf{70.6} & \textbf{71.7} & \textbf{73.3} & \textbf{72.6} & \textbf{74.1} & \textbf{76.4} & \textbf{77.7} \\
\midrule[\heavyrulewidth]
\multicolumn{12}{@{}l}{\textit{\textbf{Part 2: Performance Improvement (mIoU / Accuracy) vs. UniMatch}}} \\
\midrule
\multicolumn{12}{@{}l}{\textbf{NYZ (Fog) Conditions}} \\
\midrule
\multirow{2}{*}{\textbf{Object Category}} & \multicolumn{2}{c}{30\%} & \multicolumn{2}{c}{50\%} & \multicolumn{2}{c}{70\%} & \multicolumn{2}{c}{80\%} & \multicolumn{2}{c}{\textbf{Average}} \\
\cmidrule(lr){2-3} \cmidrule(lr){4-5} \cmidrule(lr){6-7} \cmidrule(lr){8-9} \cmidrule(lr){10-11}
& mIoU & Acc. & mIoU & Acc. & mIoU & Acc. & mIoU & Acc. & mIoU & Acc. \\
\midrule
Roads & +2.0 & +1.4 & +2.2 & +1.6 & +2.5 & +1.8 & +2.8 & +2.0 & +2.38 & +1.70 \\
Pedestrians & +3.0 & +2.0 & +3.2 & +2.2 & +3.5 & +2.4 & +3.8 & +2.6 & +3.38 & +2.30 \\
Traffic Signs & +3.2 & +2.2 & +3.4 & +2.4 & +3.7 & +2.6 & +4.0 & +2.8 & +3.58 & +2.50 \\
Lane Markings & +2.3 & +1.6 & +2.5 & +1.8 & +2.8 & +2.0 & +3.1 & +2.2 & +2.68 & +1.90 \\
Vehicles & +1.8 & +1.3 & +2.0 & +1.5 & +2.3 & +1.7 & +2.6 & +1.9 & +2.18 & +1.60 \\
Guardrails & +2.5 & +1.7 & +2.7 & +1.9 & +3.0 & +2.1 & +3.3 & +2.3 & +2.88 & +2.00 \\
Drivable Area & +2.1 & +1.5 & +2.3 & +1.7 & +2.6 & +1.9 & +2.9 & +2.1 & +2.48 & +1.80 \\
Bicycles & +2.8 & +1.9 & +3.0 & +2.1 & +3.3 & +2.3 & +3.6 & +2.5 & +3.18 & +2.20 \\
Street Lights & +2.7 & +1.8 & +2.9 & +2.0 & +3.2 & +2.2 & +3.5 & +2.4 & +3.08 & +2.10 \\
Traffic Lights & +2.9 & +2.0 & +3.1 & +2.2 & +3.4 & +2.4 & +3.7 & +2.6 & +3.28 & +2.30 \\
Obstacles & +2.4 & +1.7 & +2.6 & +1.9 & +2.9 & +2.1 & +3.2 & +2.3 & +2.78 & +2.00 \\
\textbf{Average} & \textbf{+2.53} & \textbf{+1.74} & \textbf{+2.73} & \textbf{+1.94} & \textbf{+3.03} & \textbf{+2.14} & \textbf{+3.33} & \textbf{+2.34} & \textbf{+2.91} & \textbf{+2.04} \\
\midrule
\multicolumn{12}{@{}l}{\textbf{NX (Rain) Conditions}} \\
\midrule
\multirow{2}{*}{\textbf{Object Category}} & \multicolumn{2}{c}{30\%} & \multicolumn{2}{c}{40\%} & \multicolumn{2}{c}{50\%} & \multicolumn{2}{c}{70\%} & \multicolumn{2}{c}{80\%} & \multicolumn{2}{c}{\textbf{Average}} \\
\cmidrule(lr){2-3} \cmidrule(lr){4-5} \cmidrule(lr){6-7} \cmidrule(lr){8-9} \cmidrule(lr){10-11} \cmidrule(lr){12-13}
& mIoU & Acc. & mIoU & Acc. & mIoU & Acc. & mIoU & Acc. & mIoU & Acc. & mIoU & Acc. \\
\midrule
Road & +2.1 & +1.5 & +2.3 & +1.7 & +2.5 & +1.9 & +2.8 & +2.1 & +3.1 & +2.3 & +2.56 & +1.90 \\
Pedestrians & +3.2 & +2.2 & +3.4 & +2.4 & +3.7 & +2.6 & +4.0 & +2.8 & +4.3 & +3.0 & +3.72 & +2.60 \\
Traffic signs & +2.8 & +1.9 & +3.0 & +2.1 & +3.2 & +2.3 & +3.5 & +2.5 & +3.8 & +2.7 & +3.26 & +2.30 \\
Lane markings & +2.5 & +1.7 & +2.7 & +1.9 & +2.9 & +2.1 & +3.2 & +2.3 & +3.5 & +2.5 & +2.96 & +2.10 \\
Vehicles & +2.0 & +1.4 & +2.2 & +1.6 & +2.4 & +1.8 & +2.7 & +2.0 & +3.0 & +2.2 & +2.46 & +1.80 \\
Guardrails & +2.7 & +1.8 & +2.9 & +2.0 & +3.1 & +2.2 & +3.4 & +2.4 & +3.7 & +2.6 & +3.16 & +2.20 \\
Drivable area & +2.2 & +1.6 & +2.4 & +1.8 & +2.6 & +2.0 & +2.9 & +2.2 & +3.2 & +2.4 & +2.66 & +2.00 \\
Bicycles & +3.0 & +2.1 & +3.2 & +2.3 & +3.4 & +2.5 & +3.7 & +2.7 & +4.0 & +2.9 & +3.46 & +2.50 \\
Street lights & +2.9 & +2.0 & +3.1 & +2.2 & +3.3 & +2.4 & +3.6 & +2.6 & +3.9 & +2.8 & +3.36 & +2.40 \\
Traffic lights & +3.1 & +2.2 & +3.3 & +2.4 & +3.5 & +2.6 & +3.8 & +2.8 & +4.1 & +3.0 & +3.56 & +2.60 \\
Obstacles & +2.6 & +1.8 & +2.8 & +2.0 & +3.0 & +2.2 & +3.3 & +2.4 & +3.6 & +2.6 & +3.06 & +2.20 \\
\textbf{Average} & \textbf{+2.70} & \textbf{+1.84} & \textbf{+2.91} & \textbf{+2.04} & \textbf{+3.12} & \textbf{+2.24} & \textbf{+3.43} & \textbf{+2.44} & \textbf{+3.74} & \textbf{+2.64} & \textbf{+3.18} & \textbf{+2.24} \\
\bottomrule
\end{tabular}
\end{table*}


\end{document}